\documentclass[10pt,twocolumn,letterpaper]{article}

\usepackage[pagenumbers]{cvpr} % To force page numbers, e.g. for an arXiv version

\usepackage{graphicx}
\usepackage{amsmath}
\usepackage{amssymb}
\usepackage{booktabs}
\usepackage{multirow}
\usepackage{paralist}
\usepackage[percent]{overpic}

\usepackage[pagebackref,citecolor=blue,linkcolor=blue,colorlinks,urlcolor=blue,bookmarks=false]{hyperref}

\usepackage[capitalize]{cleveref}
\crefname{section}{Sec.}{Secs.}
\Crefname{section}{Section}{Sections}
\Crefname{table}{Table}{Tables}
\crefname{table}{Tab.}{Tabs.}

\def\cvprPaperID{369} % *** Enter the CVPR Paper ID here
\def\confName{CVPR}
\def\confYear{2023}

\usepackage{color}
\definecolor{crimson}{rgb}{0.86, 0.08, 0.24}
\definecolor{gray}{rgb}{0.5,0.5,0.5}
\definecolor{green}{rgb}{0, 0.4, 0}
\definecolor{orange}{rgb}{1, 0.5, 0}
\definecolor{mahogany}{rgb}{0.75, 0.25, 0.0}
\definecolor{purple}{rgb}{0.6, 0, 0.6}
\definecolor{darkgreen}{rgb}{0, 0.4, 0}
\definecolor{frenchblue}{rgb}{0.0, 0.45, 0.73}
\definecolor{red}{rgb}{1,0,0}
\definecolor{yellow}{rgb}{1,1,0}
\definecolor{magenta}{rgb}{1,0,1}
\definecolor{pink}{rgb}{1,0.412,0.706}
\definecolor{newgreen}{rgb}{0, 0.6, 0.2}

\newcommand{\mycomment}[1]{}
\newcommand{\comment}[1]{}

\definecolor{userOurs}{rgb}{0.27, 0.45, 0.77}
\definecolor{userBLess}{rgb}{0.92, 0.26, 0.21}
\definecolor{userDFill}{rgb}{0.44, 0.68, 0.28}
\definecolor{userNS}{rgb}{0.98, 0.74, 0.02}
\definecolor{userReal}{rgb}{0.65, 0.65, 0.65}
\definecolor{userImg2stylegan}{rgb}{0.86, 0.52, 0.76}

\makeatletter
\DeclareRobustCommand\onedot{\futurelet\@let@token\@onedot}
\def\@onedot{\ifx\@let@token.\else.\null\fi\xspace}

\def\eg{\emph{e.g}\onedot} 
\def\ie{\emph{i.e}\onedot}

\makeatother

\def\eg{e.g.,~}               % for example
\def\ie{i.e.,~}               % that is, in other words
\newlength\paramargin
\newlength\figmargin
\newlength\subfigmargin
\newlength\presecmargin
\newlength\secmargin
\newlength\subsecmargin
\newlength\tabmargin
\newlength\eqmargin

\newcommand{\subsecref}[1]{Section~\ref{subsec:#1}}
\newcommand{\figref}[1]{Figure~\ref{fig:#1}} 
\newcommand{\tabref}[1]{Table~\ref{tab:#1}}

\long\def\ignorethis#1{}

\begin{document}

\title{InfiniCity: Infinite-Scale City Synthesis}

\author{
Chieh Hubert Lin$^{1,2}$ \,\, Hsin-Ying Lee$^2$ \,\, Willi Menapace$^{2,3}$  \,\, Menglei Chai$^2$  \\
Aliaksandr Siarohin$^2$ \,\, Ming-Hsuan Yang$^{1,4,5}$ \,\, 
Sergey Tulyakov$^2$ \\ [0.5em]
$^1$UC Merced, $^2$Snap Inc., $^3$Trento University, $^4$Yonsei University, $^5$Google Research \\ [0.5em]
\href{https://hubert0527.github.io/infinicity/}{https://hubert0527.github.io/infinicity/} \vspace{-0.3em}
% First Author\\
% \and
}

\twocolumn[{%
\maketitle
\renewcommand\twocolumn[1][]{#1}%
    \centering 
    \vspace{-5.0mm}
    % \begin{overpic}[width=\linewidth]{imgs/teaser_first.png}
    %     \put(60,-40){\includegraphics[scale=0.21]{imgs/maps/000005.png}}  
    % \end{overpic}
    \includegraphics[width=\linewidth]{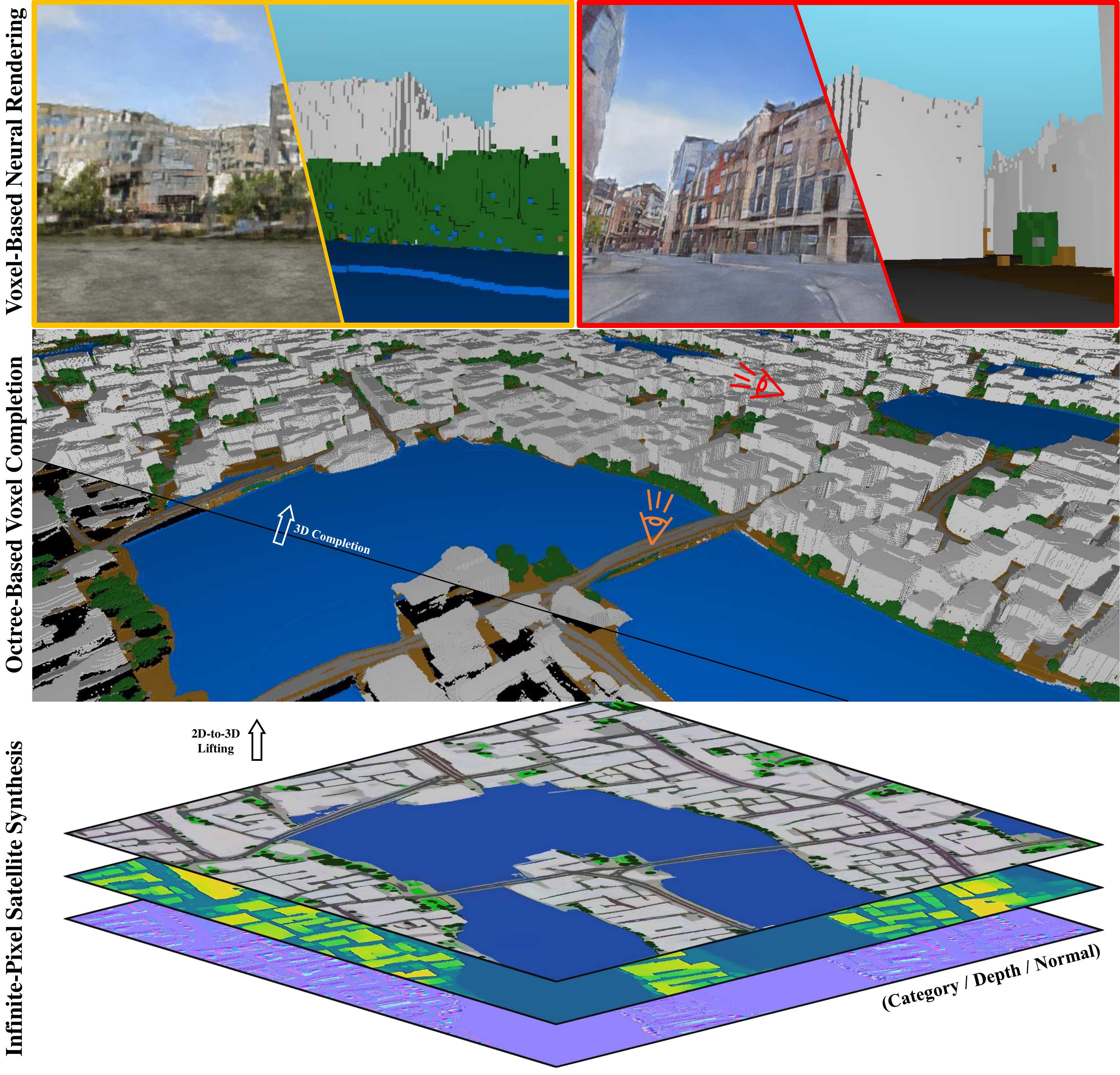}
    \vspace{-1.5em}
    \captionof{figure}{
    \textbf{We propose InfiniCity, a three-stage synthesis framework toward infinite-scale city scene synthesis.}
    Starting from the bottom to the top, we synthesize multi-modality infinite-pixel satellite images, perform octree-based voxel completion to create a watertight voxel world, then finally texturize with voxel neural rendering.
    In the middle figure, we mark the camera locations (in {\color{red}red} and {\color{orange}orange}) used to render the views in the top figures.
    % The {\color{red}red} boxes label the corresponding regions in the map.
    } \label{fig:teaser}
    \vspace{3mm}
}]

%%%%%%%%% ABSTRACT
\begin{abstract}
Toward infinite-scale 3D city synthesis, we propose a novel framework, InfiniCity,
which constructs and renders an unconstrainedly large and 3D-grounded environment from random noises.
InfiniCity decomposes the seemingly impractical task into three feasible modules, taking advantage of both 2D and 3D data.
First, an infinite-pixel image synthesis module generates arbitrary-scale 2D maps from the bird's-eye view. 
Next, an octree-based voxel completion module lifts the generated 2D map to 3D octrees.
Finally, a voxel-based neural rendering module texturizes the voxels and renders 2D images.
InfiniCity can thus synthesize arbitrary-scale and traversable 3D city environments, and allow flexible and interactive editing from users. 
We quantitatively and qualitatively demonstrate the efficacy of the proposed framework.

\end{abstract}

\section{Introduction}
\label{sec:intro}

With the rapid evolution in the generative modeling research, modern generators can synthesize high-quality images with high-fidelity~\cite{rombach2022high,dhariwal2021diffusion,Karras2019stylegan2}, 3D consistent content with neural rendering~\cite{gu2021stylenerf,chan2022efficient,hao2021gancraft}, and temporal-consistent videos~\cite{mallya2020world,ho2022video,brooks2022generating}.
However, most of these works focus on perfecting the synthesis quality with limited camera movement or within a bounded space, therefore less suitable for modeling an unconstrainedly large scene.
Recent attempts to achieve infinite visual synthesis with neural implicit model~\cite{lin2022infinitygan} or connecting anchors~\cite{skorokhodov2021aligning} are only feasible for the side view of landscapes or city scenes and create unrealistic global structures at extreme-large field-of-views, while another effort toward navigable 3D indoor scene synthesis~\cite{devries2021unconstrained} is designed for bounded environments with a constrained camera distribution and the synthesis is of limited resolution.
%Despite several recent works proposing to achieve infinite visual synthesis with neural implicit model~\cite{lin2022infinitygan} or connecting anchors~\cite{skorokhodov2021aligning}, these methods are only feasible for the side view of landscapes or city scenes and create unrealistic global structures at extreme-large field-of-views.
% 
Such limitations pique our interest in synthesizing infinite-sized, realistic, and navigable 3D environments.
% 
%Notably, GSN~\cite{} is a highly relevant approach toward navigable 3D indoor scene synthesis, however, not only its resolution is limited, GSN is designed for bounded environments with a constrained camera distribution.

% In this work, we propose InfiniCity, a pipeline that can synthesize 3D city of arbitrary scale.
Among all 3D environments, we take city scenes as a case study as they are ubiquitous in contemporary gaming, virtual reality, and augmented reality as an everyday sight.
%Aiming at synthesizing infinite-scale city scenes, we develop InfiniCity, an effective pipeline embracing the benefits from both 2D and 3D data.
%With these observations, we develop InfiniCity, a conceptually simple but effective pipeline toward infinite-scale 3D environment synthesis, and take the city scene as a case study.
% 
%\hubert{I died here...}
% 
%Instead of choosing sides between learning from 2D or 3D data, we embrace the benefits from both ends.
Ideally, it is desirable and straightforward to synthesize the whole 3D environment all at once. However, it is currently impractical with existing techniques and hardware constraints.
% 
% We observe that we can break down the pipeline into the global structure and local information generation.
We identify that the synthesis of a 3D environment can be broken down into stages of global structure planning and local perfection.
% With such observation, we propose InfiniCity, a pipeline that breaks down the whole synthesis procedure into stages of global structure planning and local perfection.
% 
First, intuitively, the satellite view of a city illustrates the outlines of the city, and provides abundant clues about the major structural information. 
As the leftover missing information remains local, we can therefore finish the structure with local 3D completion, then finalize the textures with view-dependent neural rendering.
% Therefore, we train a satellite image synthesis model in RGBD modality, which creates a surface sketch of the city.
% Then, followed by a 3D completion pipeline

With these observations, we develop InfiniCity, an effective pipeline toward infinite-scale 3D environment synthesis embracing the benefits from both 2D and 3D data.
As shown in \figref{teaser}, InfiniCity consists of three major modules.
First, we learn to generate arbitrary-scale 2D maps from bird-eye views of 3D data, using patch-based implicit image synthesis~\cite{lin2022infinitygan}.
Adopting map generation as our initial step facilitates easy user-editing, as shown in \figref{interactive}.
Then, we lift the 2D maps to 3D representation by a 3D-completion module. 
Here, we adopt the octree representation for memory and computation efficiency.
Finally, we perform neural rendering~\cite{hao2021gancraft} on the generated octrees using real 2D images as well as pseudo-realistic images generated via state-of-the-art semantic synthesis method~\cite{park2019SPADE}.

As the first attempt toward the infinite-scale generation of an unbounded environment, we conduct quantitative and qualitative experiments on the HoliCity dataset~\cite{zhou2020holicity} to corroborate the necessity of multi-stage pipeline using both 2D and 3D data, then ablate various pivotal design choices. and pivotal operations, including patch contrastive learning and bilateral filtering.

The main contributions are listed as follows:
\begin{compactitem}
\item We propose InfiniCity, an efficient and effective pipeline toward infinite-scale 3D city scene generation.
\item We demonstrate a sophisticated framework breaking down the originally improbable task into sub-modules, making the best use of both 2D and 3D data. 
\item We design an interactive sampling GUI to enable fast and flexible user interaction.
\end{compactitem}

\section{Related Work}
\label{sec:related}

\paragraph{Infinite Visual Synthesis}
Several recent attempts explore modeling infinitely large environments using only finite images with limited field-of-view.
% There are two major directions with the scope. 
One direction is to generate arbitrary-scale static images with a divide-and-conquer strategy where small patches are synthesized and then merged~\cite{lin2019coco}. 
The inference process can either be autoregressive~\cite{esser2021taming,wu2022nuwa} or non-autoregressive~\cite{lin2022infinitygan,skorokhodov2021aligning}.
Another branch of work focuses on the long-range generation of novel views along with a camera trajectory~\cite{li2022infinitenature,infinite_nature_2020}.
However, the videos rendered by such an approach lack 3D consistency, since the appearances are modeled in the 2D image space without utilizing the 3D representations.
In this work, we explore the infinite synthesis in the 3D city scene, 
aiming to generate a 3D-grounded traversable environment of infinite scale. 

It is worth noting a few concurrent works also explore unbounded 3D scene generation under different camera assumptions.
SGAM~\cite{shen2022sgam} proposes to render the surface of the 3D scene from pure satellite view.
PersistentNature~\cite{chai2023persistentnature} and SceneDreamer~\cite{chen2023sd} explores camera fly-through video similar to InfiniteNature~\cite{infinite_nature_2020}.
However, none of these works focus on the within-the-scene navigable scene generation similar to our setting.

\begin{figure*}[t!]
    \centering
    \includegraphics[width=\linewidth]{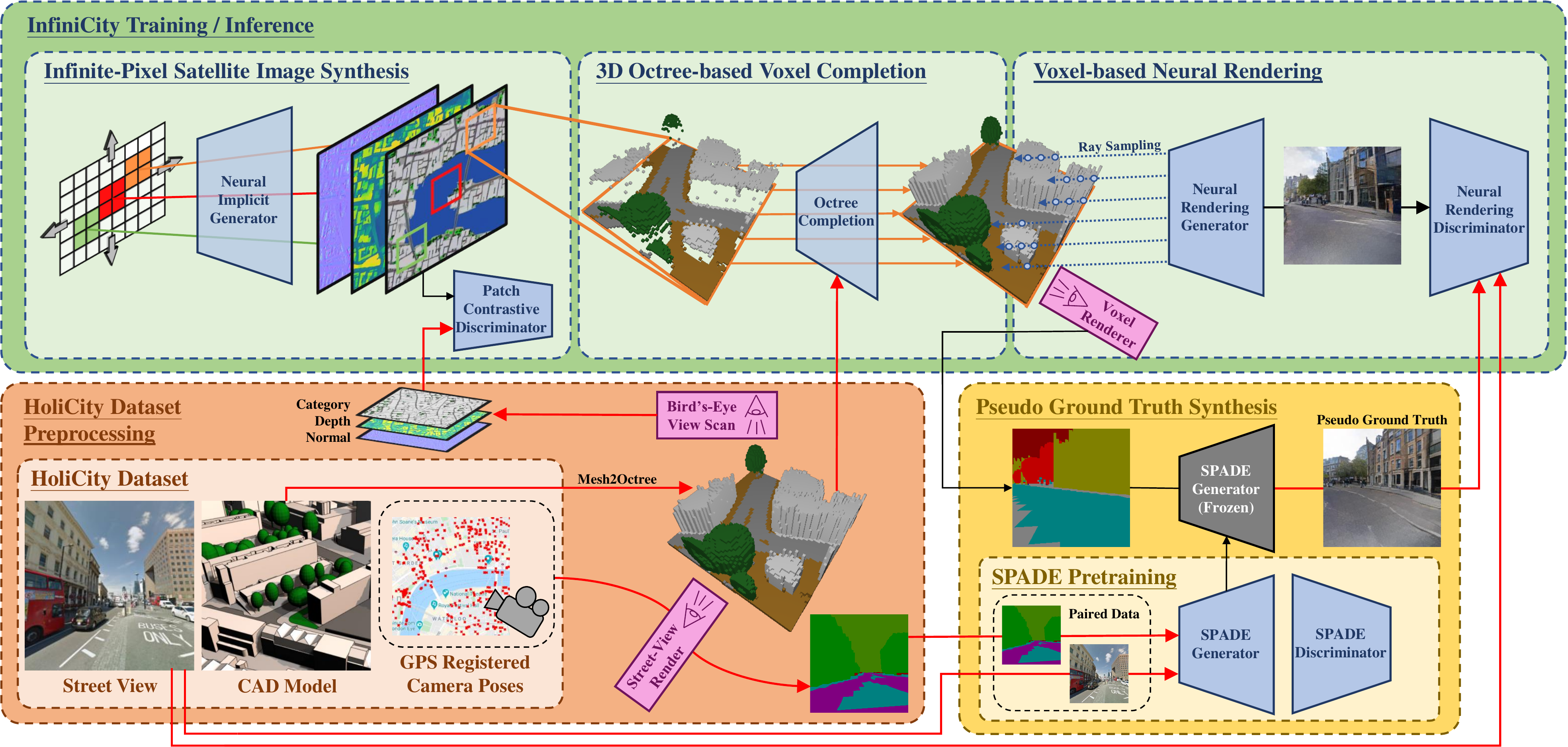}
    \vspace{-6mm}
    \caption{
    \textbf{Overview.}
    InfiniCity consists of three major modules. 
    The \textbf{\textit{Infinite-pixel satellite image synthesis}} stage is trained on image tuples (category, depth, and normal maps) derived from a bird's-eye view scan of the 3D environment, and is able to synthesize arbitrary-scale satellite maps during inference.
    The \textbf{\textit{3D octree-based voxel completion}} stage is trained on pairs of surface-scanned and completed octrees. During inference, it takes the surface voxels lifted from the satellite images as inputs and produces the watertight voxel world.
    % and performs lifting from generated satellite maps to 3D voxel world during inference. 
    Finally, the \textbf{\textit{voxel-based neural rendering}} stage performs ray-sampling to retrieve features from the voxel world, then renders the final image with a neural renderer. The neural renderer is trained with both real images and pseudo-ground-truths synthesized by a pretrained SPADE generator.
    % then perform texturization to the voxel world.
    With these modules, InfiniCity can synthesize an arbitrary-scale and traversable 3D city environment from noises. 
    }
    \vspace \figmargin
    \vspace{-1mm}
    \label{fig:pipeline}
\end{figure*} 

\paragraph{3D Generation from 3D Data}
To learn a 3D generative model, it is intuitive to train it on 3D data. 
Recent works have explored various 3D representations, including point clouds~\cite{achlioptas2018learning,luo2021diffusion}, voxels~\cite{smith2017improved,xie2018learning}, signed distance functions~\cite{chen2019learning,cheng2022sdfusion,cheng2022cross,mittal2022autosdf}, etc.
In this work, we also leverage explicit 3D supervision to learn the geometry of the 3D environment. 
Specifically, we adopt octree~\cite{riegler2017octnet,wang2017cnn,wang2018adaptive,wang2020deep}, a sparse-voxel representation, as our 3D representation.

\paragraph{3D Generation from 2D Data}
Inspired by Neural Radiance Fields (NeRF)~\cite{mildenhall2020nerf}, NeRF-based generators~\cite{chan2022efficient,chan2021pi,gu2021stylenerf,schwarz2020graf,siarohin2023unsupervised,skorokhodov20233d,skorokhodov2022epigraf,xu2023discoscene} combined with GAN-based framework~\cite{goodfellow2014generative,Karras2019stylegan2} have become a dominating direction to learn 3D structure from 2D image collections.
These methods are restricted to modeling single objects within a limited range of viewing angles from a camera. 
Relaxing the constraints, GSN~\cite{devries2021unconstrained}  models traversable indoor scenes with many local radiance fields, yet it requires trajectories of calibrated images for training and is not directly applicable to city scenes due to the unbounded nature and complexity of outdoor scenes.
The other branch of work targets performing texturization on a given 3D representation. %which differs from our exploration of creating the environment from scratch with both structure and texture synthesis. 
Given a 3D object~\cite{siddiqui2022texturify}, a 3D scene~\cite{jeong20223d}, or a voxel world~\cite{hao2021gancraft}, colorization and texturization are learned with differentiable rendering followed by a GAN-based framework.

\vspace{-2.5mm}
\section{InfiniCity}
\label{sec:method}
\vspace{-1.5mm}
We aim to generate infinite-scale 3D city scenes using both 2D and 3D data.
In Figure~\ref{fig:pipeline}, the InfiniCity synthesis pipeline consists of three main components.
The infinite 2D-map synthesis first generates an arbitrarily large satellite map from random noises, the octree-based voxel completion model converts the map into a watertight voxel environment, then neural rendering texturizes the voxel world.
We further discuss each of the components.

\vspace{-1mm}
\subsection{Data Preprocessing}
\vspace{-2mm}
The dataset consists of images with GPS-registered camera poses ${I, p}$ and CAD model $C$ representing the scenes.
We further process the data for each of the three modules. 
For the octree-based voxel completion, we first convert and partition the CAD model to a set of octrees $\{O_i\}$, each representing a sub-region of the city. 
These octrees are not only the supervision for the 3D completion training but also further render into two different types of training data.% with ``bird's-eye view scan'' and ``street-view render.''
The ``bird's-eye view scan'' extracts multiple modalities of the octree surface information into surface octrees $\{O_i^\mathrm{sur}\}$ from the top-down direction, then these surface octrees are further converted into 2D images
(\ie categorical, depth, and normal maps), jointly denoted as $I^{\textrm{CDN}}$.
The $\{O_i^\mathrm{sur}\}$ pairing with $\{O_i\}$ constitutes the 3D completion model training data, while $I^{\textrm{CDN}}$ serves as the training data for the infinite-pixel satellite image synthesis.
On the other hand, the ``street-view render'' utilizes the GPS-registered camera location along with annotated camera orientation ${p_j}$ to render segmentation images ${I^\textrm{seg}_j}$ corresponding to the street-view images ${I_j}$. % provided by the HoliCity dataset.
Such procedure constructs data pairs for training SPADE~\cite{park2019SPADE}, which later synthesizes the pseudo-ground-truth during the neural rendering training.

\vspace{-1.5mm}
\subsection{Infinite 2D Map Synthesis}
\label{subsec:infinity}
\vspace{-1.5mm}
% 1. Contrastive patch-D
% 2. Force Depth alignment?
% 3. Color-based categorical synthesis
% 4. Bilateral Filtering
Infinite-scale generation directly on 3D data is currently far-fetched, yet recently explored on 2D images.
Therefore, instead of directly generating the 3D environment, we propose to start by synthesizing the corresponding 2D map.
We leverage the infinite-pixel image synthesis ability of InfinityGAN~\cite{lin2022infinitygan}, which synthesizes arbitrarily large maps with the neural implicit representation. 
Due to the limitations of  data, we generate categorical labels instead of real RGB satellite images.
However, as GANs encounter problems in propagating gradients while modeling discrete data~\cite{yu2017seqgan,zhang2017adversarial}, we instead  assign colors to each of the classes and train InfinityGAN on the categorical satellite map rendered with the assigned colors.
The colors are later converted back to a discrete category map with the nearest color.
Meanwhile, to convert the predicted satellite images to a voxel surface for the next-stage 3D shape completion, we jointly model the height map information.
To further regularize the structural plausibility, we further model the surface normal vector, which is the aggregated average surface normal over the unit region covered by a pixel in the satellite view.

In InfiniCity, a critical problem is that the satellite image has a larger field-of-view, compared to the original InfinityGAN setting.
Accordingly, InfiniCity requires extra focus on the structural plausibility in the local region.
Directly applying adversarial learning on a large and dense matrix makes the discriminator focus on global consistency and overall visual quality, instead of the local details.
Therefore, we apply the contrastive patch discriminator~\cite{park2020swapping} to increase the importance of the fine-grained details.

We synthesize tuples of images of arbitrary scale in this stage: $\hat{I}^{\textrm{CDN}} = G_{\infty}(z)$,
where all the inputs and outputs of $G_{\infty}(\cdot)$ can be of arbitrary spatial dimensions.

% \vspace{-1mm}
\subsection{Voxel World Completion}
\label{subsec:voxel-completion}
\vspace{-1.5mm}
% 1. Octree-based approach
% 2. Octree blend
The voxel world completion stage aims to create a watertight 3D representation from the synthesized maps $\hat{I}^{\textrm{CDN}}$,
% % Hubert: this should be \hat{O}, and it is not defined yet
% from the previously generated surface voxels ${O^{\mathrm{sur}}}$, 
as the following neural rendering involves ray-casting and requires reasonable ray-box intersection in the 3D space.
A critical issue of voxel representation is its immense memory consumption, due to allocating unnecessary memory to the unused empty spaces.
We utilize an octree-based voxel representation~\cite{wang2020deep,wang2017ocnn} to avoid the memory issue, and implement the 3D completion model with O-CNN~\cite{wang2017ocnn} framework for efficient neural operations directly on octrees.
%
%We construct the octrees with the categorical labels and surface normal vectors predicted from the satellite image synthesis stage.
% 
To retain the surface information, we build skip connections using OUNet~\cite{wang2020deep}, trained with voxels of spatial size $64^3$.
The model is trained with the paired data $\{O_i, O_i^\mathrm{sur}\}$.
At inference time, we partition the maps $\hat{I}^{\textrm{CDN}}$ generated in the previous stage into patches of $64^2$ pixels,  convert them into surface voxels in octree representation $\hat{O}_i^\mathrm{sur}$, and obtain 3D-completed voxels via $\hat{O}_i = G_\mathrm{vox}(\hat{O}_i^\mathrm{sur})$ for each patch. 
As a spatially contiguous city surface is already illustrated by the satellite view, we empirically observe that the separately processed octree blocks remain contiguous after the 3D completion and subsequent spatial concatenation. %spatially concatenated again.

%\paragraph{Bilateral filtering.} 
Despite the outputs being already visually plausible while using the raw output from the satellite-image synthesis step, we can still observe some isolated pixels caused by artifacts generated in the depth channel.
These artifacts create floating voxels after converting the satellite image to the surface voxels, and further lead to undesirable structures after applying the 3D completion model.
We employ bilateral filters~\cite{tomasi1998bilateral} to suppress these noises.
The filter is applied multiple times with different space and color thresholds. We first apply large kernels with small color thresholds to create sharper edges for the buildings, then small kernels with large color thresholds to remove isolated pixels.
% , as shown in \figref{bilateral}.
% 
%In particular, the tree category has its own special characteristics, it has a round shape with spatially varying depth, therefore we do not apply the bilateral filters to this particular data category.

\vspace{-1mm}
\subsection{Texturization via Neural Rendering}
\label{subsec:neural}
\vspace{-1.5mm}
% 1. Camera distribution
% 2. Walkable region labeling
% 2. Faster convergence on trajectory-based importance sampling
Finally, with a 3D-completed and watertight voxel environment, we can cast rays and utilize neural rendering to produce the final appearance of the realistic 3D environment.
% are one step away from a realistic 3D environment.
Here, we render the texturized images using  GANcraft~\cite{hao2021gancraft}, a state-of-the-art neural rendering framework that has shown success in large-scale outdoor scene synthesis.
Following its training paradigm, we first train a SPADE~\cite{park2019SPADE} model with paired segmentation and realistic street-view images $\{I^\textrm{seg}_j, I_j\}$,
and use it to create pseudo-ground-truths $\{I^\mathrm{pseudo}_k\}$ given segmentation maps sampled by the street-view renderer using random camera poses $p_k$.
The watertight octrees patches $\{\hat{O}_i\}$ are concatenated and converted into the GANcraft parametrized voxel representation $\hat{V}$, where each of the voxels is parameterized by the parameters attached to its eight corners. 
Then, for each of the \textit{valid} camera views, we cast view rays and extract the per-pixel trilinear-interpolated features base on the ray-box intersection coordinates in the 3D space.
The neural-rendering model $G_\mathrm{neural}$ is trained with randomly paired real images $I_j$ and pseudo images $I^\mathrm{pseudo}_k$ produced with camera poses $\{p_k\}$, and renders the synthesized images $\{\hat{I}_k\}$ based on the features retrieved with $\{p_k\}$.

However, a critical issue arises when it comes to sampling the \textit{valid} $\{p_k\}$.
To match the training distribution of the SPADE model, which provides the important pseudo-ground-truth for each camera view, we sample the camera near the ground instead of the simple fly-through camera used in GANcraft.
Such a deviation leads to another issue.
A city scene is occupied by lots of buildings and trees, thus it is important to detect unwanted collisions between the camera and objects.
In practice, we manually select several \textit{walkable} classes (\eg road, terrain, bridge, and greenspace), and label these voxels from the satellite view.
As SPADE has poor performance with low entropy inputs (\eg directly facing a wall with a uniform class), we apply three steps of erosion and connected component labeling to remove the small alleys and the small squares hidden between buildings.
We sample the camera locations with such labeled zone, along with randomly sampled camera orientations\footnote{Based on the statistics used in HoliCity~\cite{zhou2020holicity}, we sample arbitrary yaw and roll, and $0\deg-45\deg$ pitch.}.
We also observe the training becomes less stable and sometimes produces spiking gradients, thus an R1 regularization~\cite{mescheder2018training} is added to stabilize the training.

In summary, we can formulate the overall infinite-scale 3D scene generation process as:
\vspace{-1mm}
\begin{equation*}
    \small
    \begin{aligned}
       &\text{\subsecref{infinity}: } \, \hat{I}^{\textrm{CDN}} = G_{\infty}(z),\\
        &\text{\subsecref{voxel-completion}: } \, \{\hat{O}_i^\mathrm{sur}\} = \mathrm{convert}(\hat{I}^{\textrm{CDN}}), 
        \{\hat{O}_i\} = G_\mathrm{vox}(\{\hat{O}_i^\mathrm{sur}\}), \\
        &\text{\subsecref{neural}: } \, V = \mathrm{aggregate}(\{\hat{O}_i\}),
        \{\hat{I}_k\} = G_\mathrm{neural}( V, \{p_k\}).\\
    \end{aligned}
    \vspace{-1mm}
\end{equation*}

Within the final texturized 3D environment $V$, we can traverse and visualize scenes given desired camera trajectories ${p_i}$ via neural rendering.

\begin{figure}[t]
    \centering
    \includegraphics[width=\linewidth]{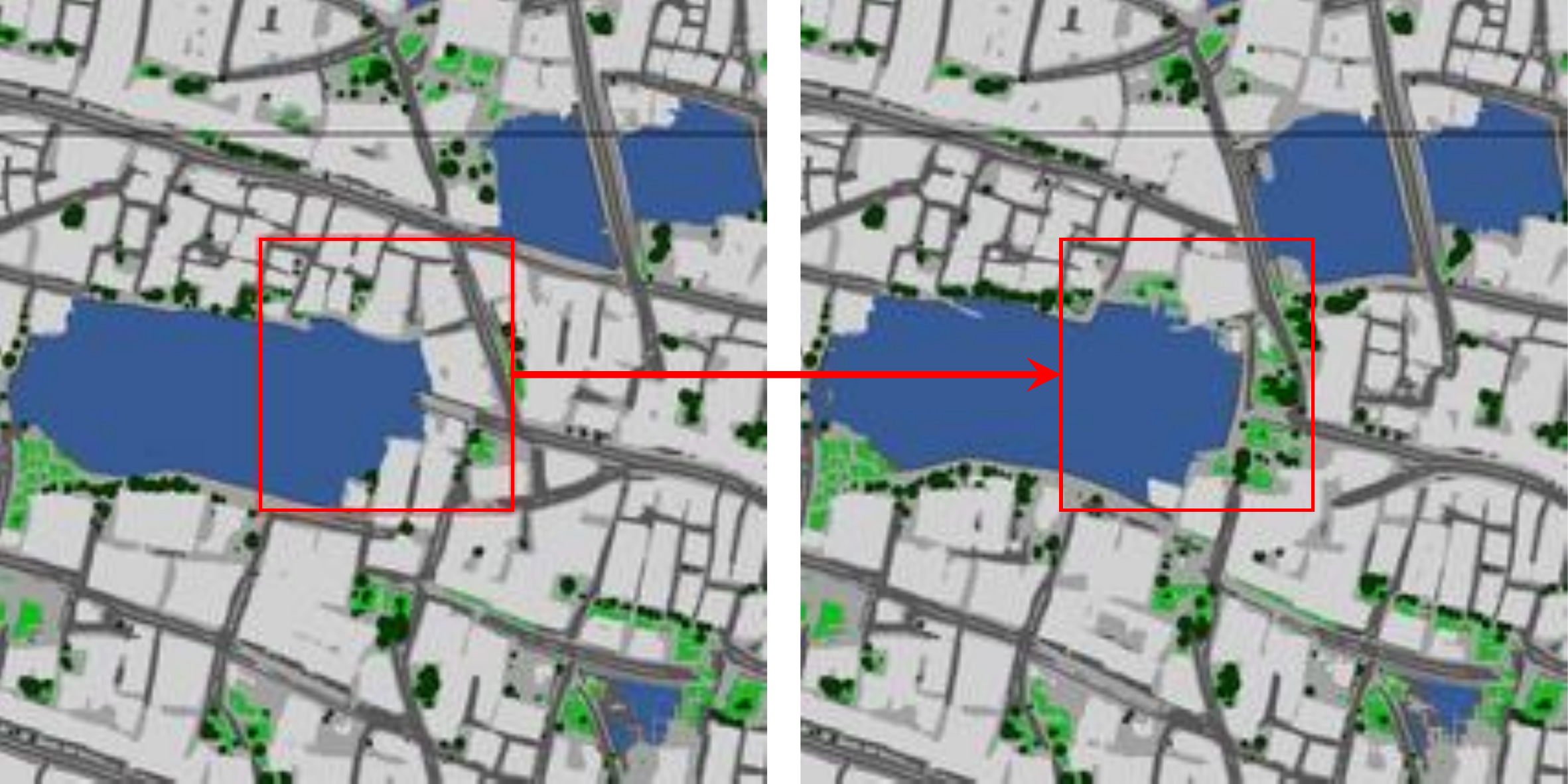} 
    \vspace{-1.7em}
    \caption{
    \textbf{Interactive resampling.}
    Our GUI allows users to select a region of interest and resample the local variables with efficient on-demand inference operated only on the neighbor regions. 
    Notice that an undesired "the road running into the lake" results is alleviated.
    % 
    % We show the complete GUI in Appendix.
    }
    % \vspace \figmargin
    \vspace{-3mm}
    \label{fig:interactive}
\end{figure} 

% \vspace{-1mm}
\subsection{Interactive Sampling GUI}
\vspace{-1.5mm}
Despite InfinityGAN automating the map synthesis, it is difficult for generative models to maintain universally high-quality results over the whole map.
The difficulty of maintaining a consistent quality grows exponentially as the spatial size of the image increases.
Furthermore, certain artifacts are inevitable, such as synthesizing a bridge longer than the momentary receptive field of the generator, the bridge suddenly terminates in the middle of the water as the local latent vectors corresponding to the bridge features move beyond the momentary receptive field.
To resolve the issues and enable flexible interaction, we develop an interactive sampling GUI, which resamples local latent variables and randomized noises based on user control, giving the imperfect images a second chance.
As shown in Figure~\ref{fig:interactive}, the user can select the desired region to resample.
All the latent variables in InfinityGAN are available for resampling.

In particular, to boost interactiveness, we develop a sophisticated queuing system for efficient inference.
InfinityGAN has demonstrated its ability in ``spatially independent generation'', where the generator can independently generate the image patches without accessing the whole set of local latent variables.
Such a characteristic further enables us to queue each patch synthesis task as a job in a FIFO queue, and run inference in a batch manner by tensor-stacking multiple jobs in the queue.
Furthermore, when a user selects a region of interest that intends to resample, we implement a feature calibration mechanism to collect only the subset of variables that has any contribution to the pixels within the selected region, then the subset of variables are resampled and pushed to the FIFO queue.
As such, we perform only the necessary computations with the maximum GPU utilization rate, increasing the inference speed by a large margin.
For instance, while interacting with a 4096$\times$4096 map, the original synthesis method takes 16 seconds on a single RTX2080 Ti. 
In the case of selecting a 256$\times$256 region, our mechanism takes only 1.7 seconds for locally resampling latent variables, which is a 10 times speed up.
% , and 1.4 seconds for resampling randomized noises

\vspace{-1mm}
\section{Experiments}
\label{sec:experiment}
\vspace{-1mm}

\begin{figure*}[t!]
    \centering
    \includegraphics[width=.162\linewidth]{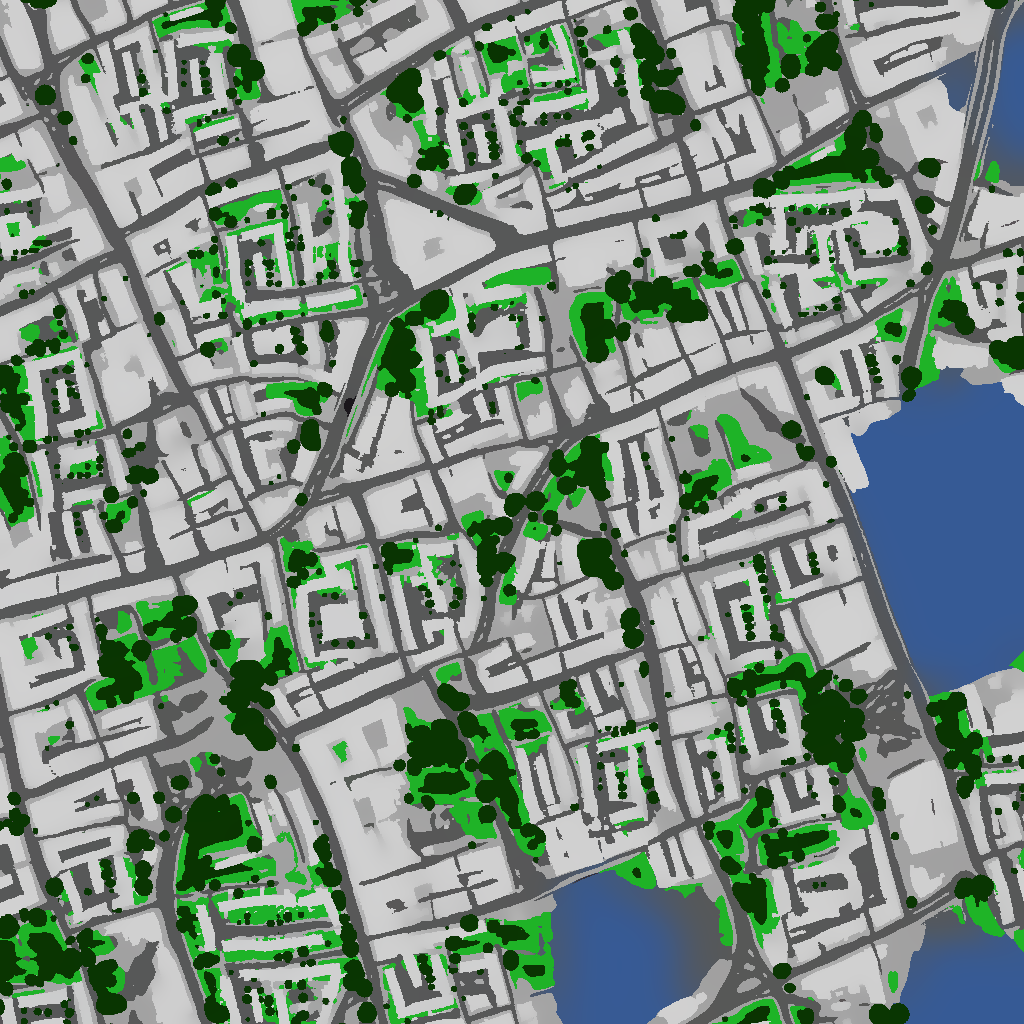}
    \includegraphics[width=.162\linewidth]{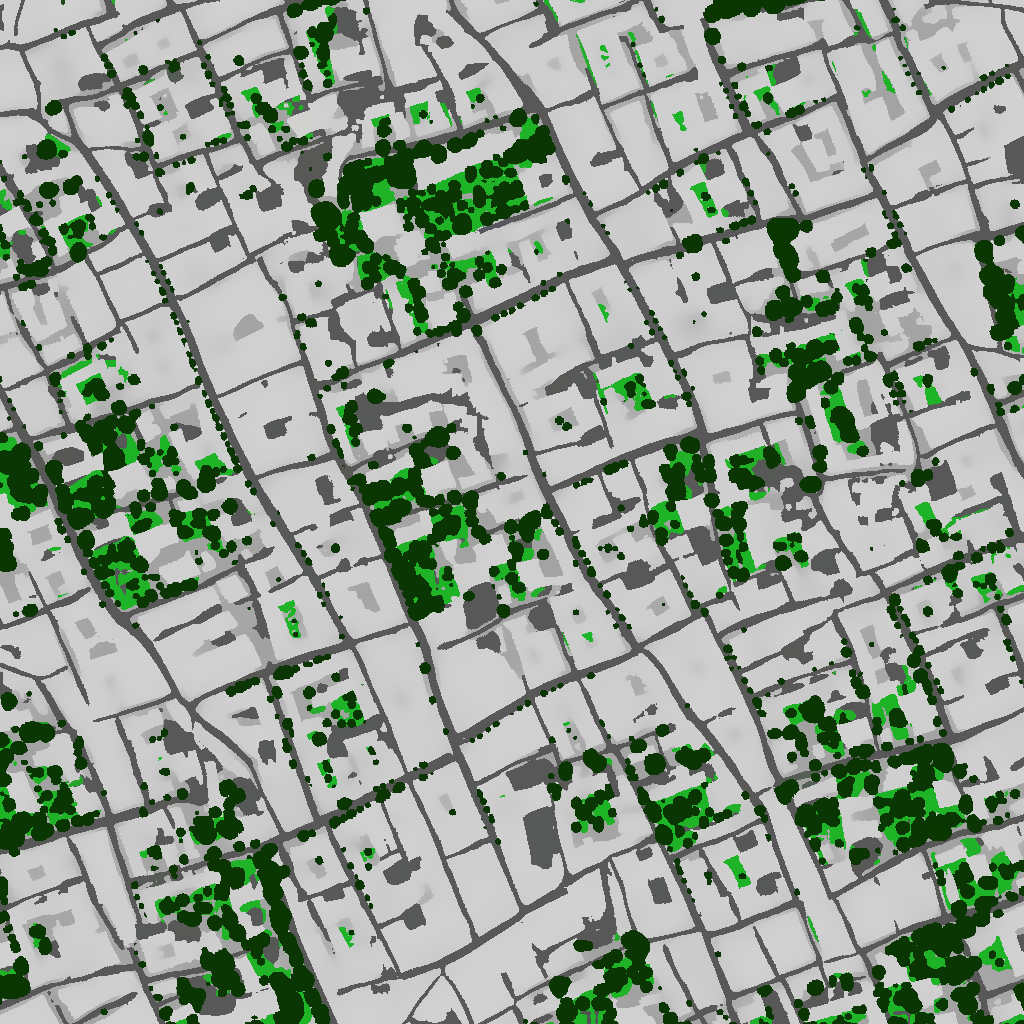}
    \includegraphics[width=.162\linewidth]{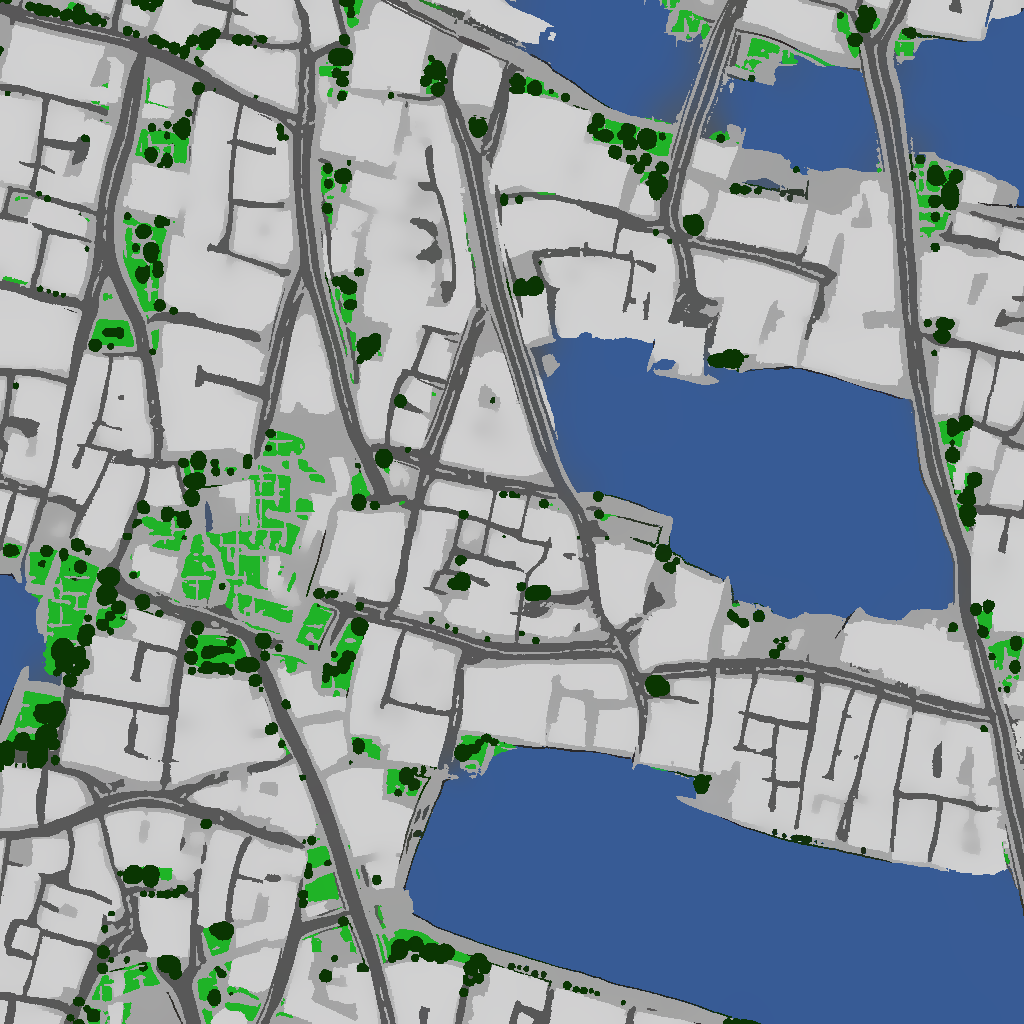}
    \includegraphics[width=.162\linewidth]{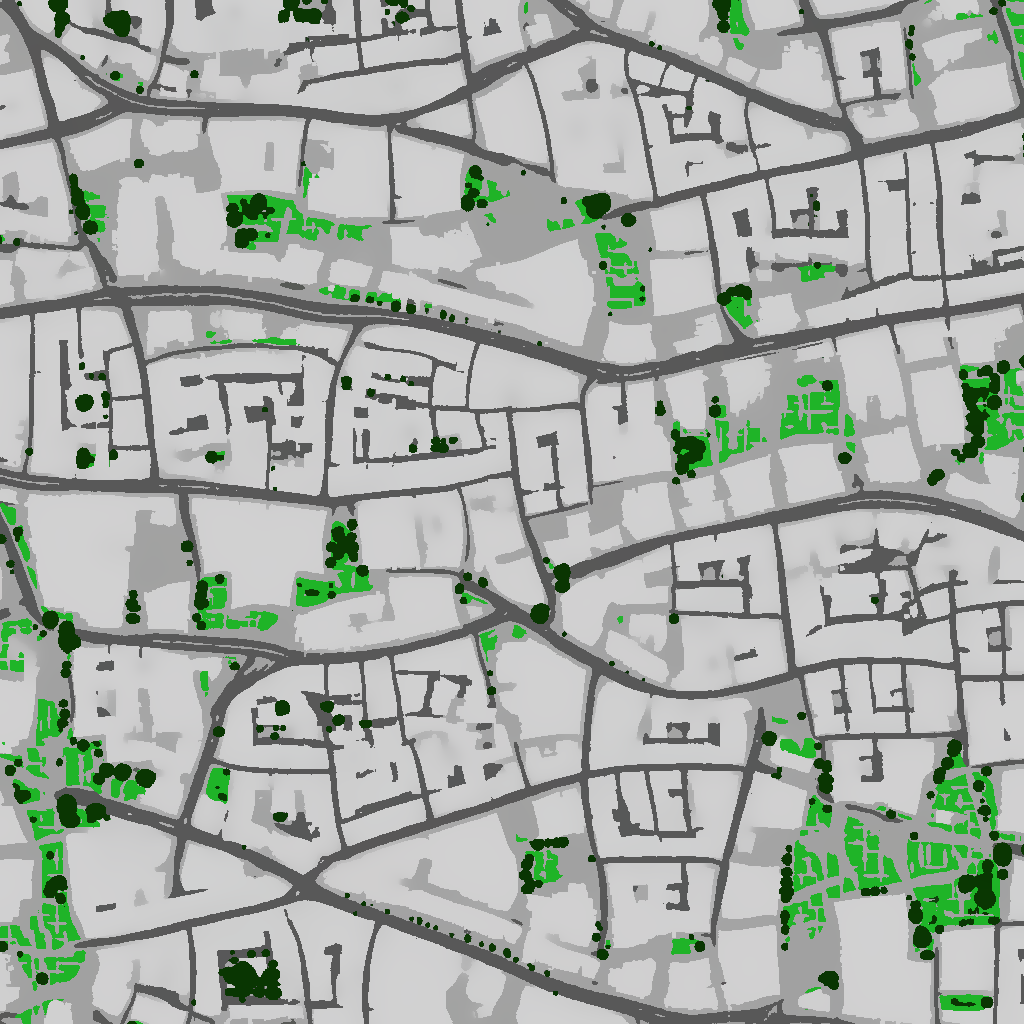}
    \includegraphics[width=.162\linewidth]{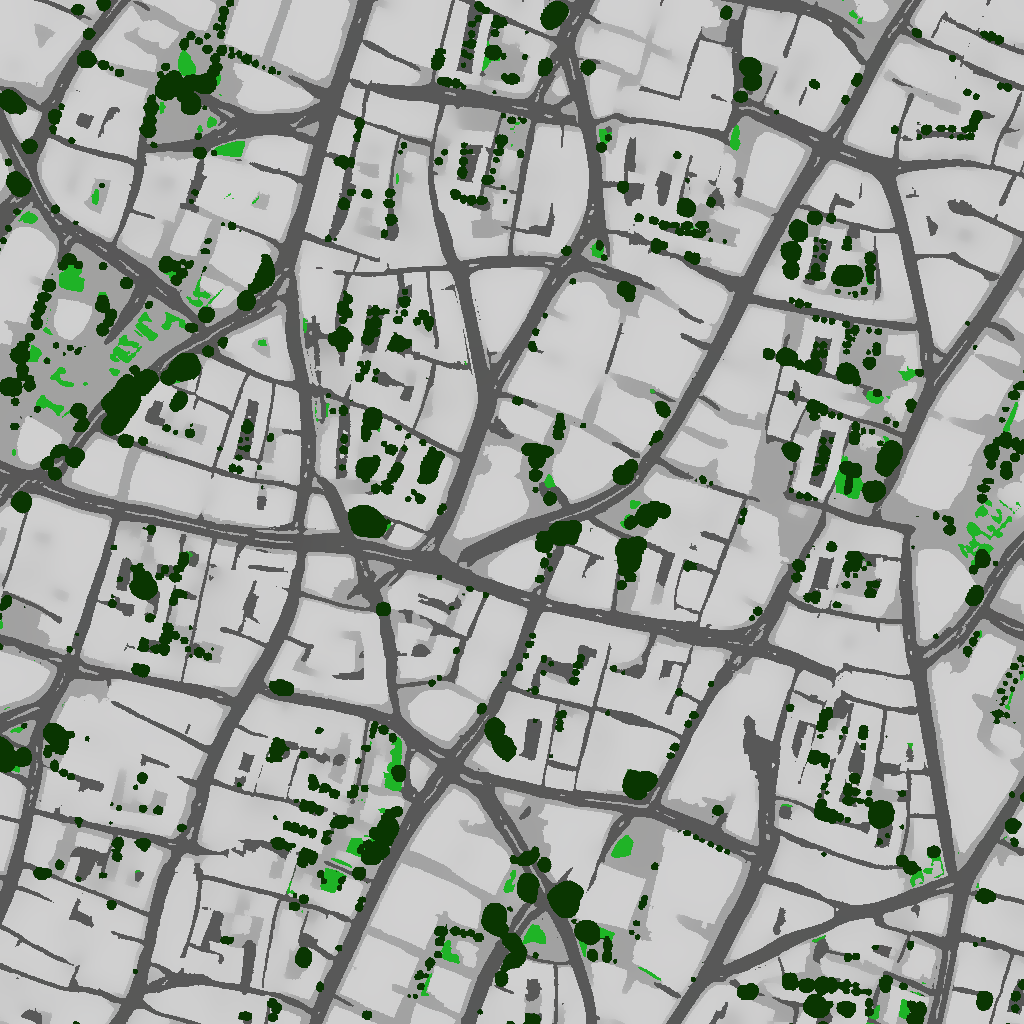}
    \includegraphics[width=.162\linewidth]{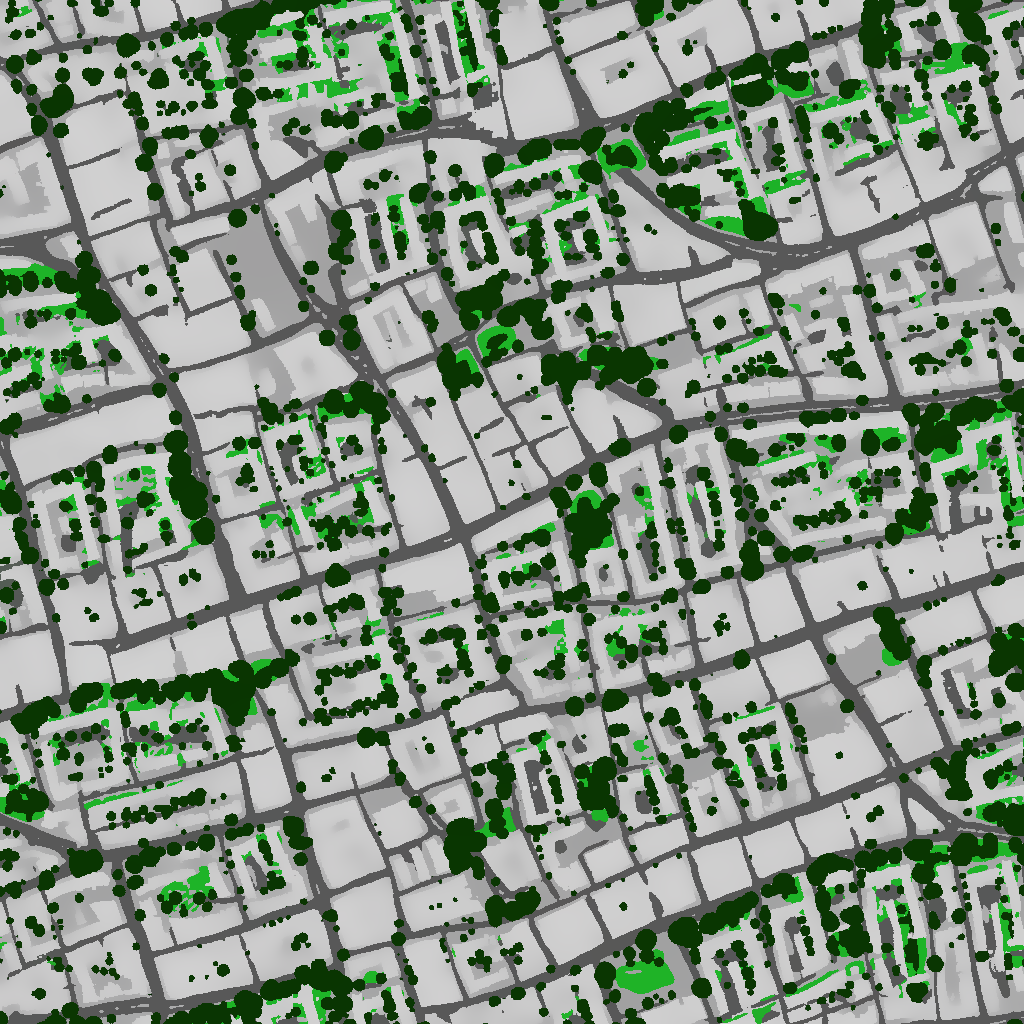}
    \\
    \includegraphics[width=.162\linewidth]{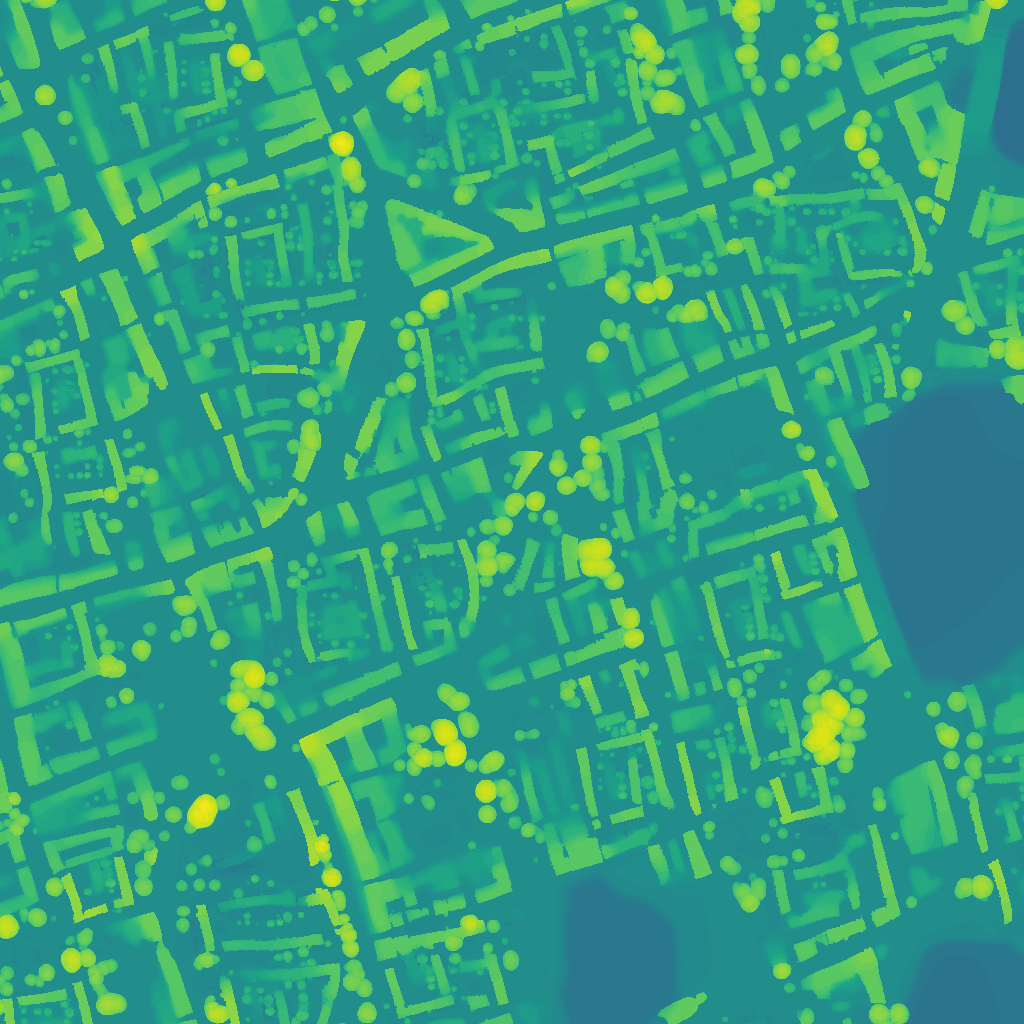}
    \includegraphics[width=.162\linewidth]{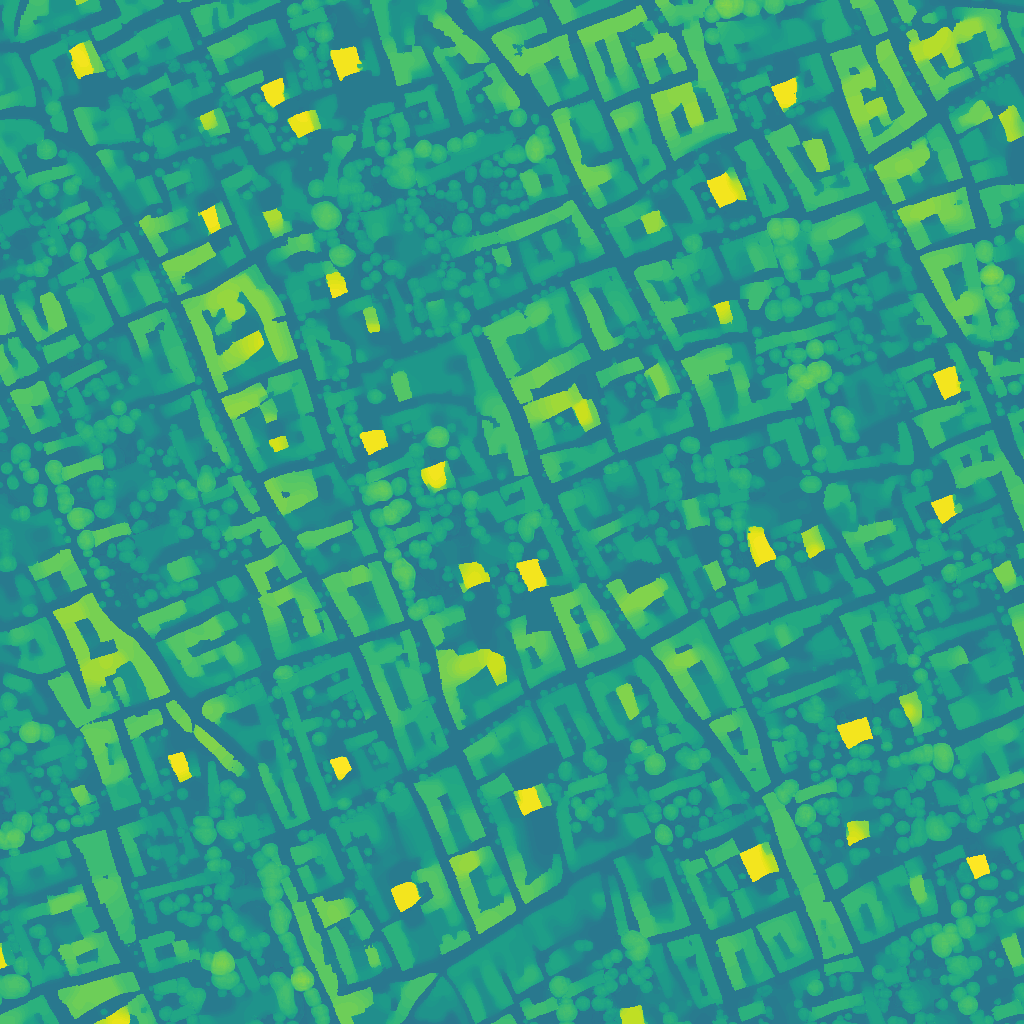}
    \includegraphics[width=.162\linewidth]{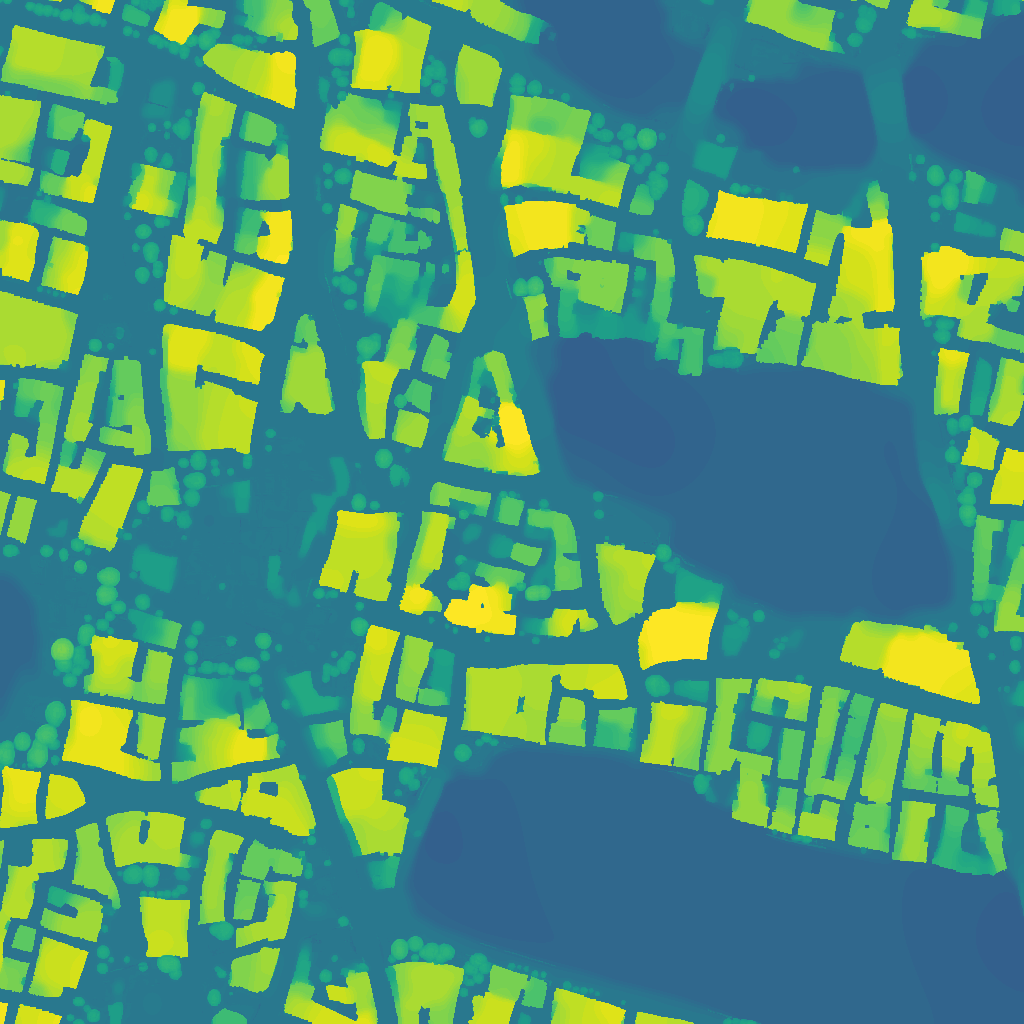}
    \includegraphics[width=.162\linewidth]{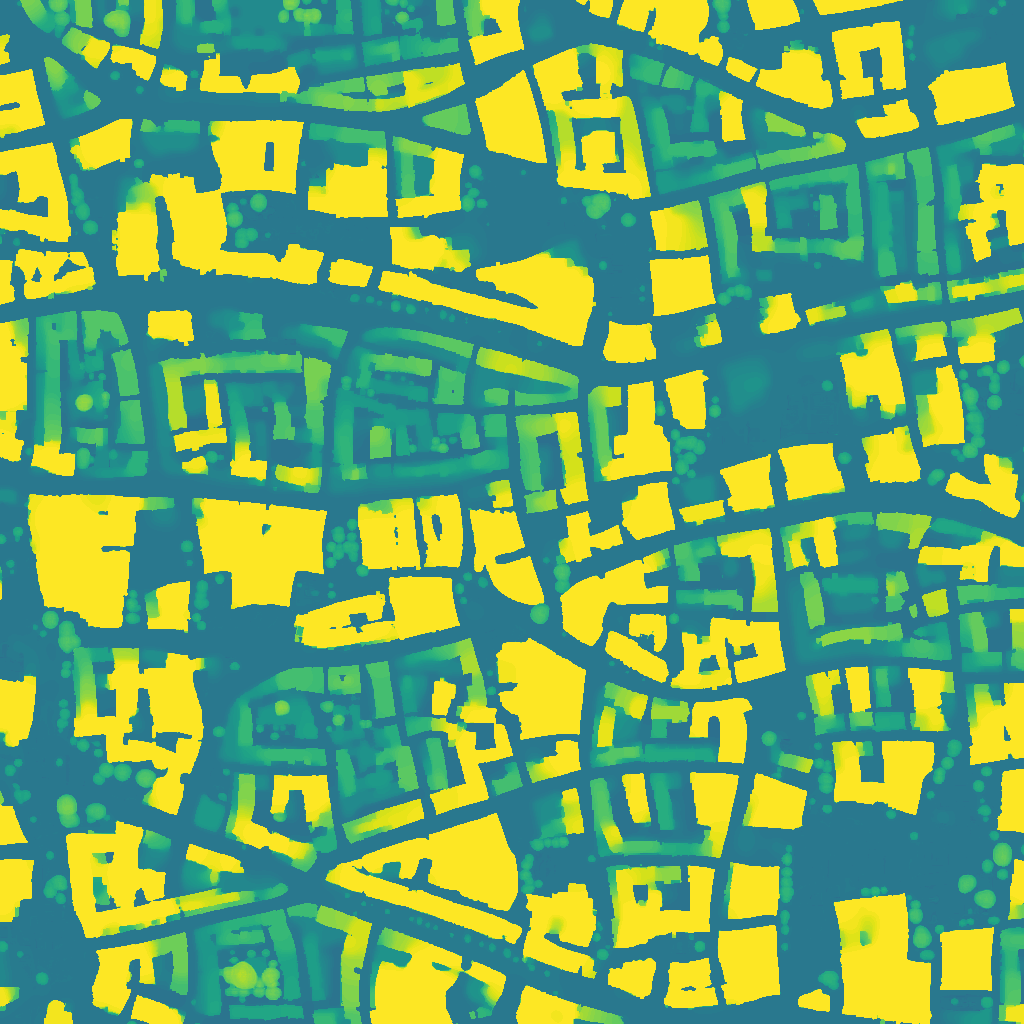}
    \includegraphics[width=.162\linewidth]{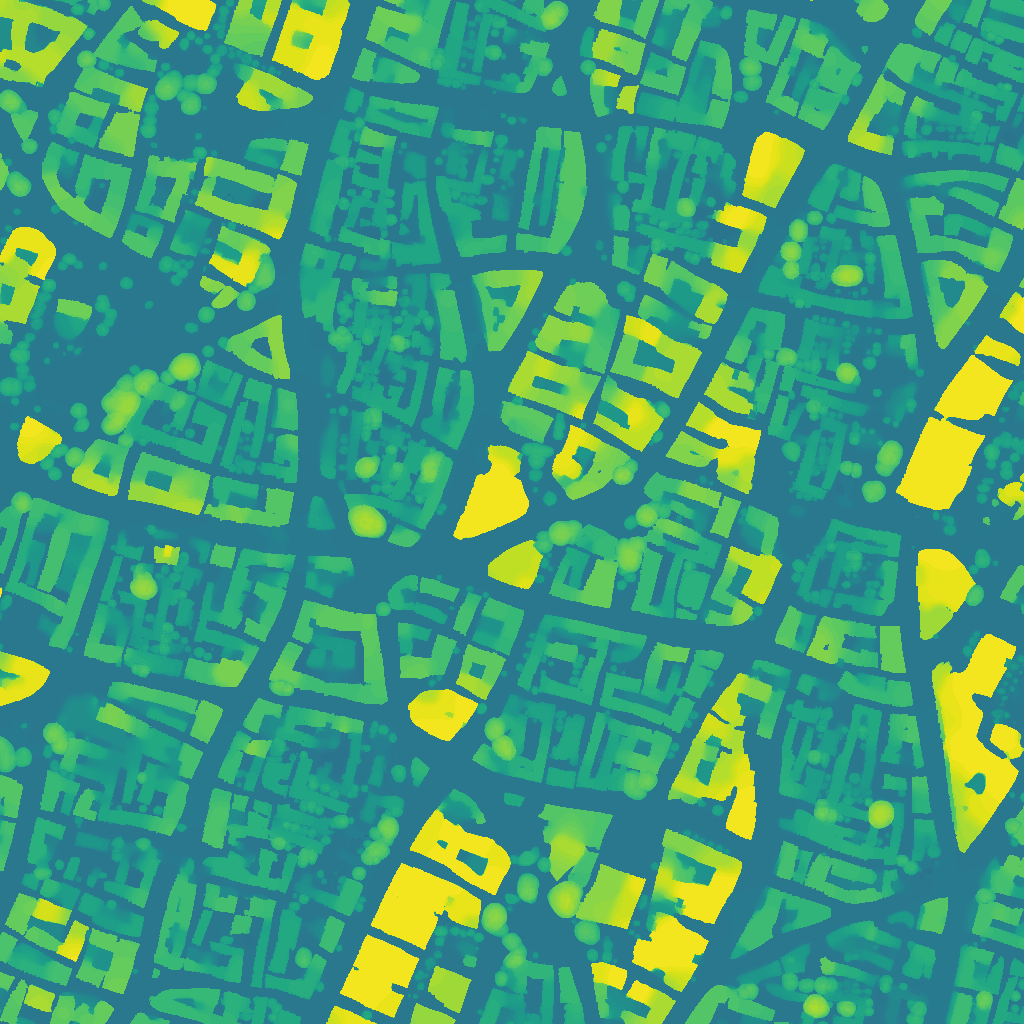}
    \includegraphics[width=.162\linewidth]{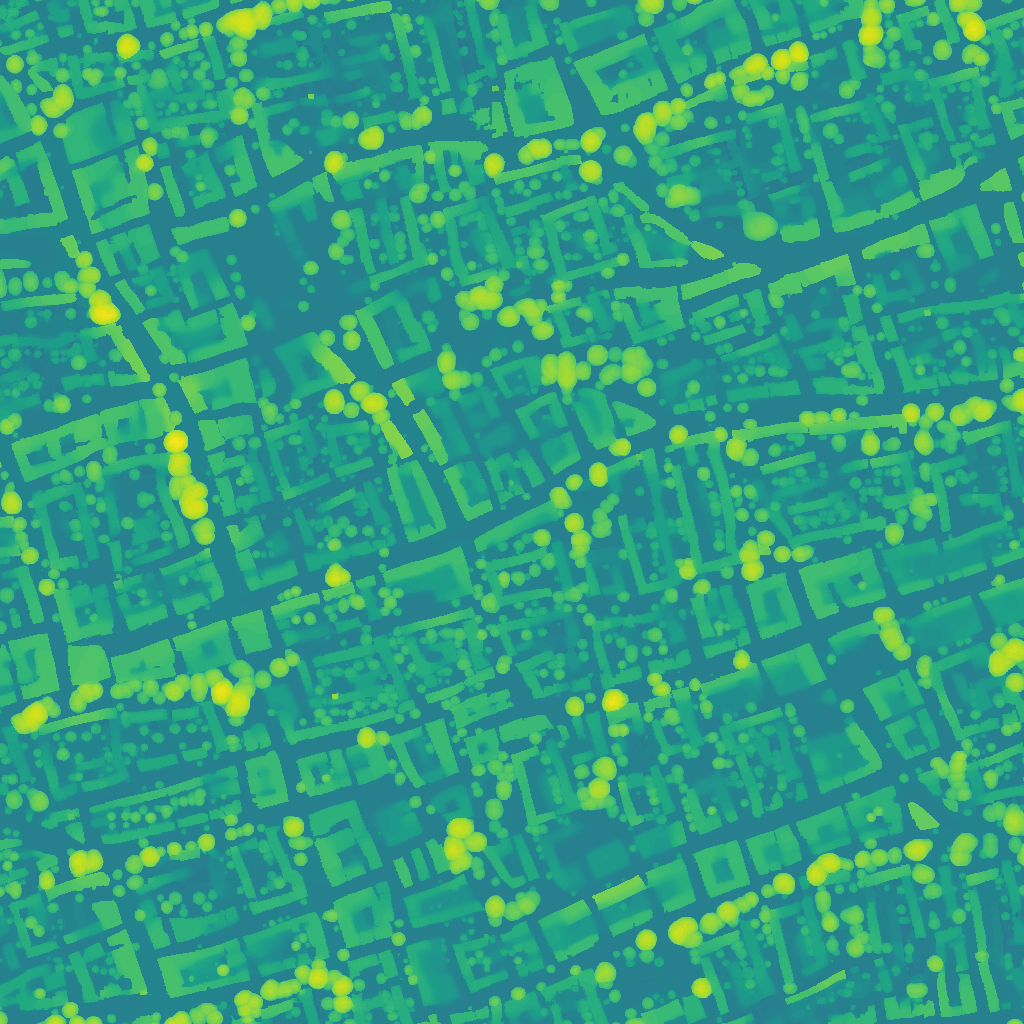}
    \\
    \includegraphics[width=.162\linewidth]{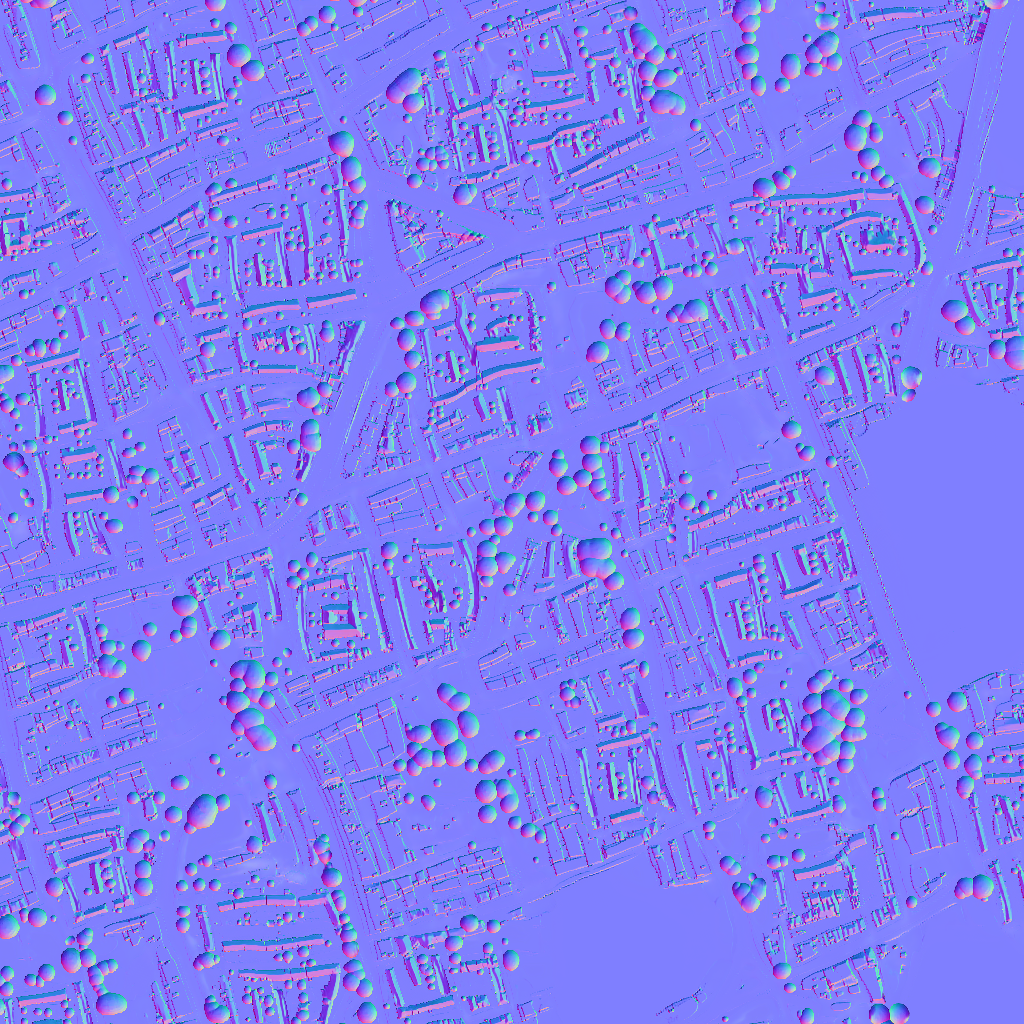}
    \includegraphics[width=.162\linewidth]{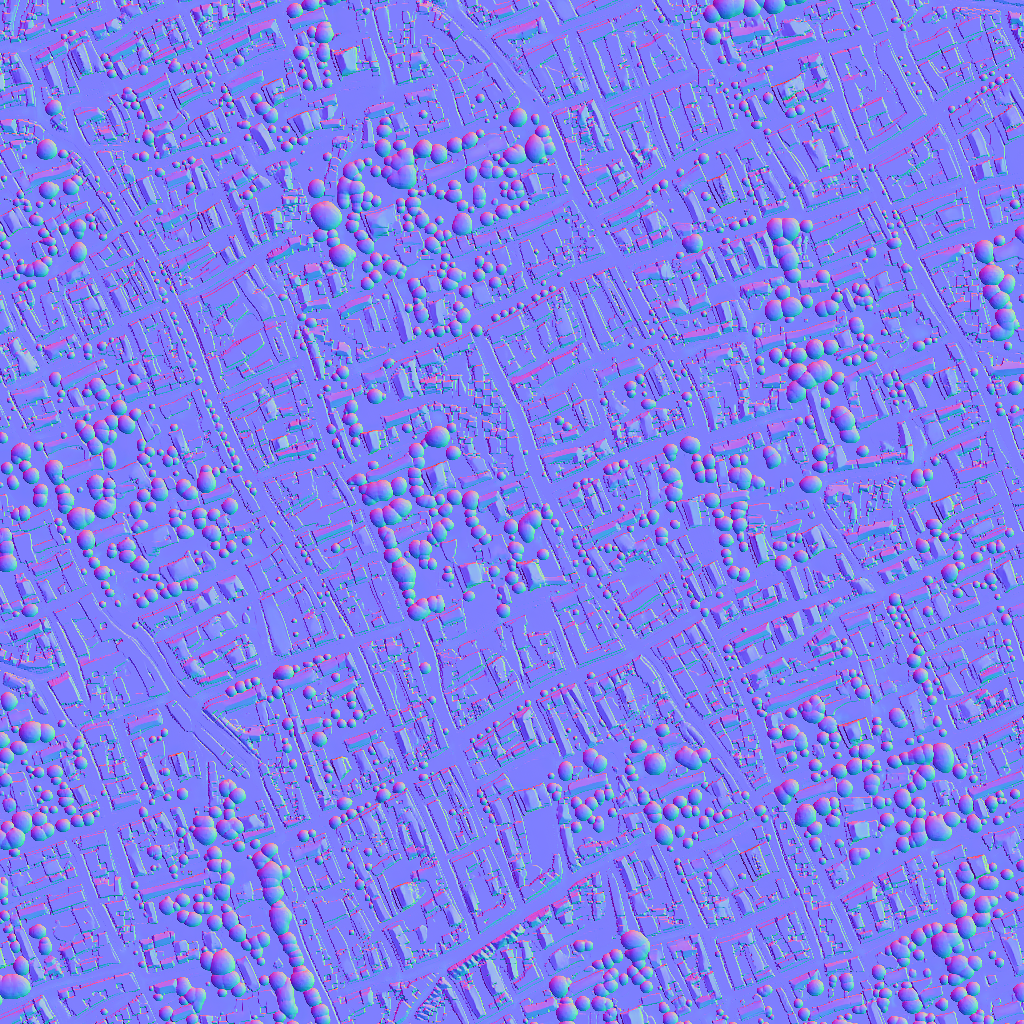}
    \includegraphics[width=.162\linewidth]{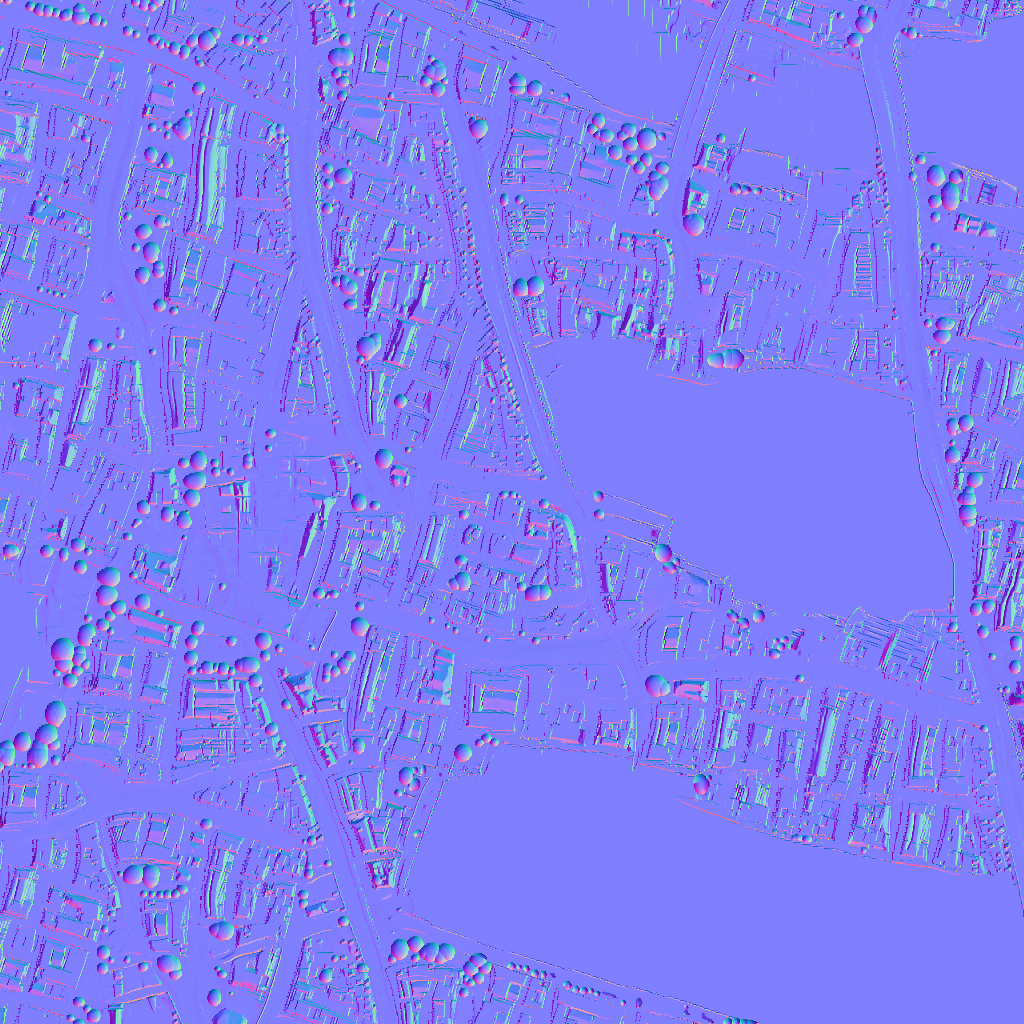}
    \includegraphics[width=.162\linewidth]{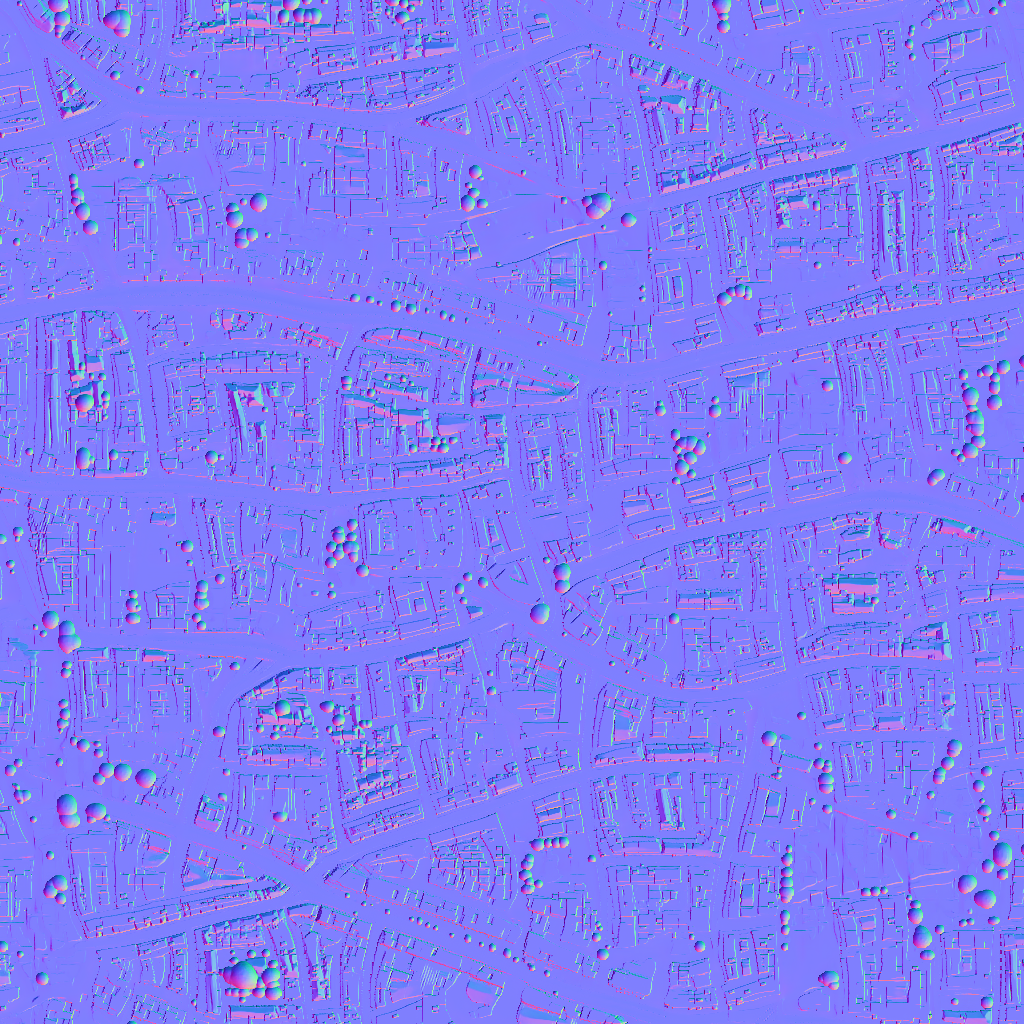}
    \includegraphics[width=.162\linewidth]{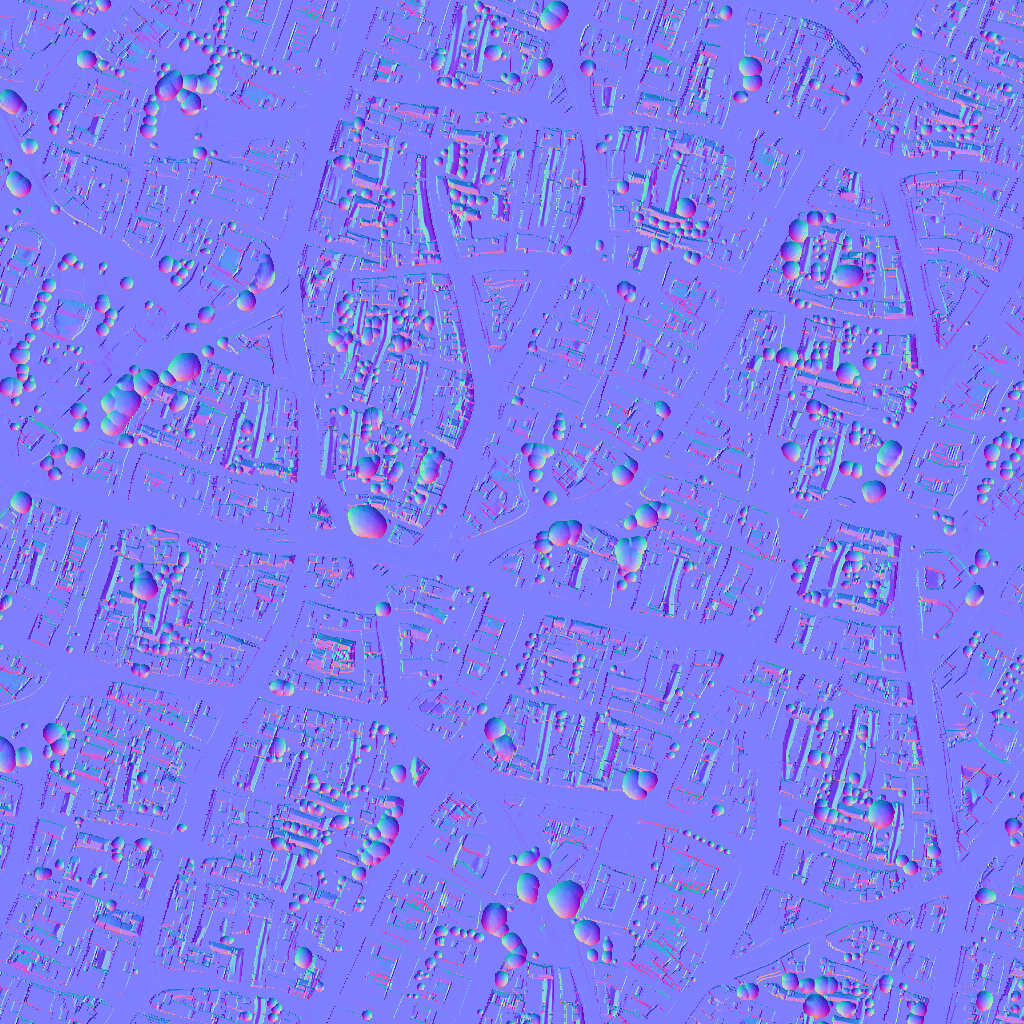}
    \includegraphics[width=.162\linewidth]{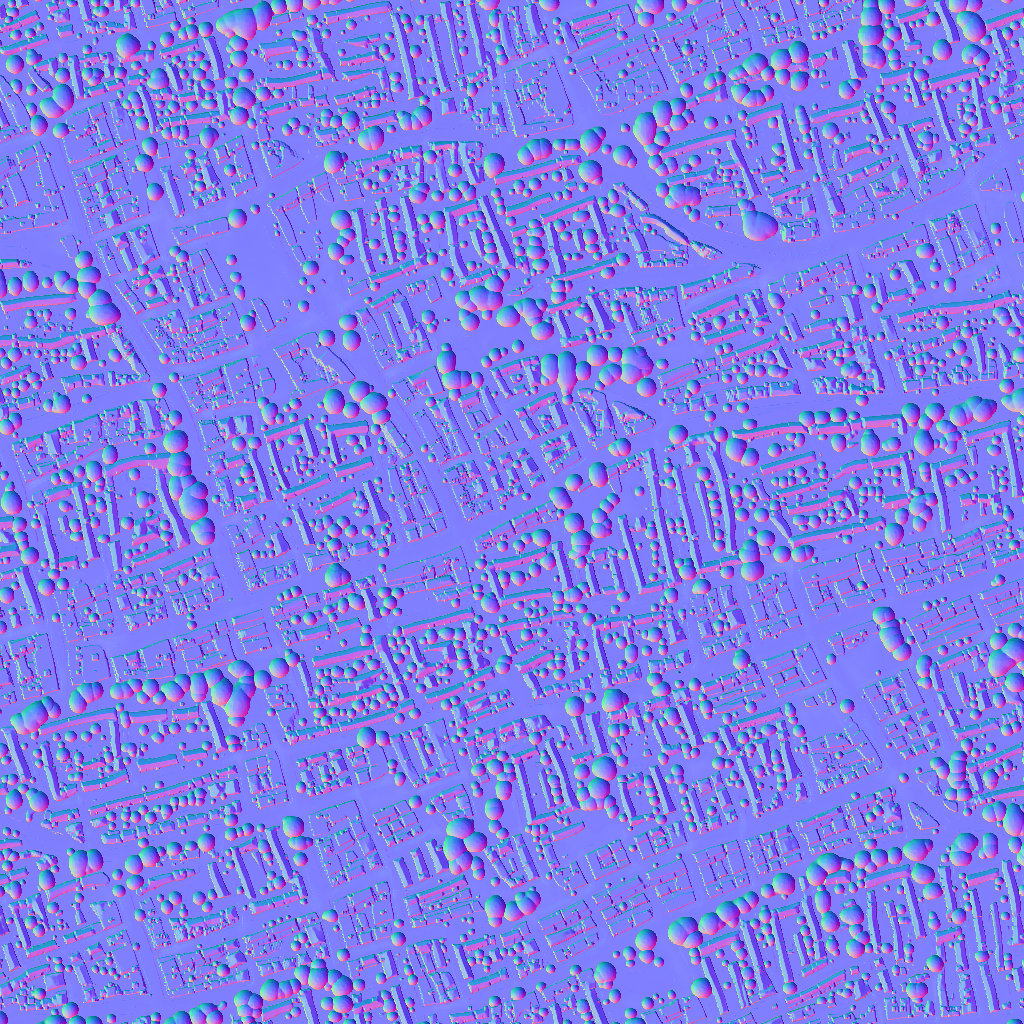} \\
    \vspace{-2mm}
    \caption{
    \textbf{Synthesized satellite maps.} 
    We train InfinityGAN with contrastive discriminator in multiple data modalities (category, depth, and normal). 
    %
    % The generator is trained at 197$\times$197 pixels, and 
    The demonstrated images are 1024$\times$1024 pixels cropped from the infinite-pixel images.
    }
    % \vspace \figmargin
    \vspace{-2mm}
    \label{fig:largemap}
\end{figure*}

\subsection{Dataset Processing}
\vspace{-1.5mm}
% 1. What is HoliCity
% 2. Mesh to points to voxels
% 3. Dataset hyperparameters: We choose voxel size 1 meter and octree depth 6
With the aforementioned InfiniCity algorithm, here we discuss how to extract the corresponding data modalities.
HoliCity~\cite{zhou2020holicity} is a large-scale dataset based on the 3D London CAD model (with object-level category annotations) collected by Accucities~\cite{accucities} and Google street-view images.
The dataset contains 50{,}024 images registered to the CAD model with GPS location along with camera orientations.
Due to the limited data accessibility, we obtain a subset of the CAD model at the L2 level of details, and the region corresponds to 14{,}612 registered images.
We use this subset to train and evaluate our algorithm.

We first perform point sampling on the meshes surface at four times voxel resolution (\ie sampling 4$^3$=64 points in each voxel grid on average).
Points are sampled at an equal distance, thus the nearest neighboring points will be within $1/4$ voxel size.
Then the space is partitioned into equal-spacing voxels. We choose the voxel resolution of one cubic meter per voxel, such a hyperparameter can be easily modified to a higher resolution with sufficient memory~\footnote{We run the InfinityGAN with 2$\times$ V100-16GB at batch size 32, OUNet with 1$\times$ V100-16GB at batch size 16, and GANcraft with 8$\times$ V100-32GB at batch size 8. Therefore, the memory consumption of InfiniCity training is mainly bounded by the neural rendering algorithm.}.
Each voxel aggregates the point information with a majority vote for the category and vector averaging for the surface normal.
The voxels are then partitioned and converted to a depth-$6$ octree, with 64 (=$2^6$) voxels on each edge.
% 
%In practice, a deeper octree with higher resolution is feasible, we leave the exploration of the appropriate resolutions as future work.
% 
To further retrieve the satellite-image training data for the infinite-pixel satellite-view synthesis module, we scan the voxels from the top-down direction, retrieve the aggregated voxel information (\ie category, depth, and normal) from the first-hit voxels, and project them onto a 2D image.

\begin{table}[th!]
    \centering
    \caption{
    \textbf{Contrastive patch learning improves InfinityGAN.}
    We train InfinityGAN at 197$\times$197 pixels. The contrastive patch discriminator shows a significant performance boost. 
    % 
    %PD is the patch discriminator, CD is the contrastive patch discriminator.
    }
    \vspace{-2mm}
    \small
    \label{tab:infinitygan}
    \begin{tabular}{l|r}
    \toprule
    Model            & FID \\
    \midrule
    InfinityGAN      & 115.6 \\
    InfinityGAN + patch discriminator & 103.8 \\ % Actually, not fully trained...
    InfinityGAN + contrastive patch discriminator &  \textbf{77.0}\\
    \bottomrule
    \end{tabular}
    \vspace{-1mm}
\end{table}

\begin{table}[th!]
\caption{
    \textbf{Quantitative comparisons in structure synthesis.}
 InfiniCity outperforms the end-to-end synthesis method PVD~\cite{Zhou_2021_ICCV}, showing the efficacy of modeling structures from the satellite view.
    % 
    % The ablations support that our design choices improve the quality of synthesized voxels.
}
\vspace{\tabmargin}
\centering
\small
\begin{tabular}{l c} 
    \toprule
    %\multirow{2}{*}{Method} & \multicolumn{2}{l}{$1 \times 1$} &    \multicolumn{2}{l}{$5\times 5$}\\
     Method & P-FID  \\
    \midrule
    PVD~\cite{Zhou_2021_ICCV}              & 12.06 \\
    InfiniCity (Ours) w/ pillar completion & 8.61 \\
    InfiniCity (Ours) w/o bilateral        & 6.92 \\
    InfiniCity (Ours)                      & \textbf{6.08}\\
    \bottomrule
\end{tabular}
% \hfill
% \begin{tabular}{l c c} 
%     \small
%     \toprule
%     Method & Precision & Recall\\
%     \midrule
%     Pillar            & 0.916 & 0.901 \\
%     Completion (Ours) & 0.882 & 0.826 \\
%     \bottomrule
% \end{tabular}
% \vspace{\tabmargin}
\vspace{-1mm}
\label{tab:pcd}
\end{table}

\begin{table}[t!]
\caption{
    \textbf{Quantitative comparison with GSN}.
    InfiniCity substantially outperforms GSN in FID and KID (both the lower the better).
}
\vspace{\tabmargin}
\centering
\small
\begin{tabular}{l r r} 
    \toprule
    Method & FID $(\downarrow)$ & KID $(\downarrow)$\\
    \midrule
    GSN~\cite{devries2021unconstrained}    & 333.92 & 0.325 $\pm$ 0.002 \\
    Ours   & \textbf{108.47} & \textbf{0.084 $\pm$ 0.001} \\
    \bottomrule
\end{tabular}
% \vspace{\tabmargin}
\label{tab:fid}
\end{table}

\begin{figure*}[t!]
    \centering
    % silly placeholders
    \includegraphics[width=0.95\linewidth]{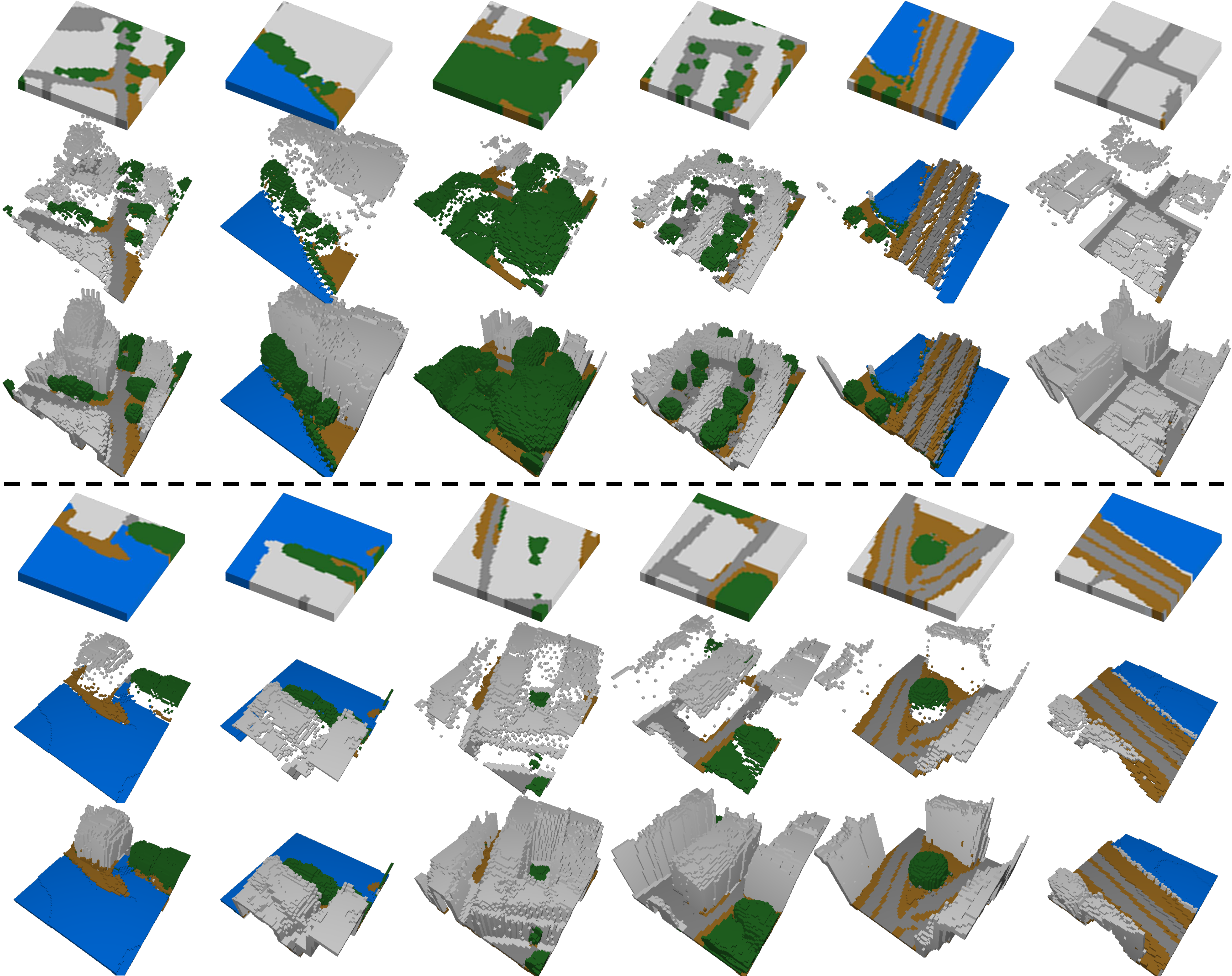}
    \vspace{-0.5em}
    \caption{
    \textbf{Octree-based voxel completion.}
    High-quality and high-diversity voxels completed from synthetic satellite images.
    % The voxels completed from synthetic satellite images maintain a high quality and high diversity.
    For both sample groups, we show synthesized satellite images, lifted surface voxels, then 3D-completed voxels.
    % The odd rows are surface voxels converted from the multi-modality satellite images, while the even rows are 3D completed voxels. 
    % 
    The samples are 64$^3$ voxels.
    % From top to bottom are: surface voxels converted from the satellite modality, 3D-completed voxels, and the extracted walkable zones.
    % 
    %\hubert{Should've rendered in orthographic camera, but no time to deal with this.}
    }
    \vspace \figmargin
    % \vspace{-0.5mm}
    \label{fig:completion}
\end{figure*} 

% \vspace{-1mm}
\subsection{Infinite 2D Map Generation}
\vspace{-1mm}
\paragraph{Multi-modality map synthesis.} 
In Figure~\ref{fig:largemap}, we show the satellite images synthesized with our method across the categorical, depth, and normal modalities.
The images are 1024$\times$1024 pixels cropped from the infinite-pixel images, which is equivalent to 1024$\times$1024 square meters in the real world.
The results show excellent structural alignments across modalities while maintaining high-quality and high-diversity appearances.

\paragraph{User editing interface.} In Figure~\ref{fig:interactive}, we show that a user can interactively resample the local latent variables within a sub-region of the map while retaining the majority of the map untouched.
In the example, the edit avoids the road from running into the river.
Note that the contents outside the red bounding box can still be slightly altered, since the latent variables on the edge of the bounding box affect content outside the bounding box based on the receptive field of the generator.
The critical local latent variables that contribute to the undesired pixels can locate anywhere within the bounding box.
One can modify the mechanism to update only the subset of the local latent variables that do not make any contributions to the pixels outside the bounding box, at the risk of potentially missing the critical local latent variables intended to be resampled.

\vspace{-4mm}
\paragraph{Ablate contrastive patch learning.} 
In Table~\ref{tab:infinitygan}, we show that the contrastive patch discriminator is an important component in the pipeline that significantly improves the generator quality.
Naively applying an additional patch discriminator does not bring a similar level of performance gain.

\vspace{-0.5mm}
\subsection{Voxel World Completion}
\vspace{-1.5mm}
In Figure~\ref{fig:completion}, we show the qualitative performance of our voxel completion model. The model ensures the final voxel structure is watertight and maintains the original voxel surfaces generated in the satellite-image synthesis step.

First, we would like to corroborate the significance of synthesizing structure from the satellite image.
Despite not being easily scalable to an unconstrainedly large environment synthesis, it is intuitive to learn an end-to-end generator that directly synthesizes the 3D environment from scratch.
We adopt PVD~\cite{Zhou_2021_ICCV}, a state-of-the-art point cloud generator, as a critical baseline.
We convert the 64$^3$ voxels into point clouds using the center coordinate of the voxels.
The model is trained end-to-end using 8{,}192 points per sample. 
Unlike the FID metric for images, there is no existing general and non-reference-based quality evaluation metric for 3D data. 
To measure the distribution distance similar to FID, we first pretrain an autoencoder using FoldingNet~\cite{yang2018foldingnet} on the real point cloud data and take the encoder as a feature extractor.
Then we compute the feature distance in the FID manner.
We denote the metric as P-FID. 
To compare with PVD, we also pick the center coordinate of the voxels to create point clouds from our synthesized voxels. 
As shown in \tabref{pcd}, the proposed pipeline outperforms PVD even though the evaluation setting is in PVD's favor (\ie it is trained and evaluated on point clouds).

\begin{figure*}[t!]
    \centering
    \includegraphics[width=\linewidth]{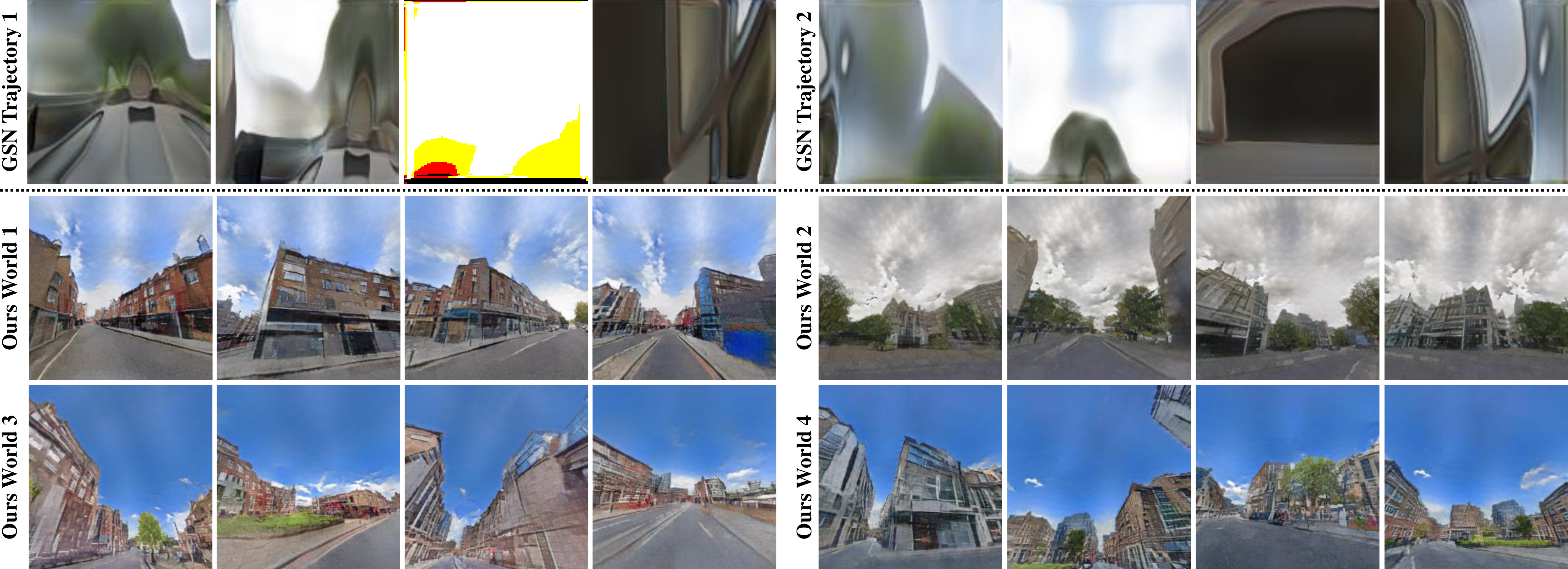}
    \vspace{-1.5em}
    \caption{
    \textbf{Trajectory-wise image rendering results.} 
    Our final rendering results present better quality, structural consistency, and diversity, over the competing method GSN~\cite{devries2021unconstrained}.
    Each group of images is rendered within the same voxel world using a shared global style code, while GSN shares the same global latent vector in each group.
    }
    \label{fig:texturization}
    \vspace{-1em}
\end{figure*} 

Next, we justify the necessity of the 3D completion modules.
A straightforward approach toward closing the gap between the surface and the ground is, for each surface voxel, to project the voxel to the ground level, and mark all voxels along the trajectory with the category of the original surface voxel.
We call this naive baseline the ``Pillar'' method, as it essentially creates a pillar for each surface voxel.
As an utterly simplified baseline method, such an approach can easily create undesired appearances for certain object classes, such as the trees.
However, we show that even such an approach, heavily borrowing the structure constructed from the satellite images, can outperform PVD, the end-to-end diffusion-based point cloud synthesis baseline.
Such results further show that synthesizing the structure from the satellite view can significantly simplify and benefit the structure synthesis procedure.
% 
% We hypothesize that the satellite view, with a sparse number of points with a significantly larger receptive field, benefits the model for synthesizing the structure.

Finally, we ablate the usefulness of bilateral filtering.
%\paragraph{Significance of bilateral filtering.}
We utilize bilateral filtering to improve the plausibility of structure and suppress the noises from the synthesized satellite depth.
In Table~\ref{tab:pcd}, we accordingly show that using bilateral filtering can further improve the P-FID.

\begin{figure}[t]
    \centering
    \includegraphics[width=\linewidth]{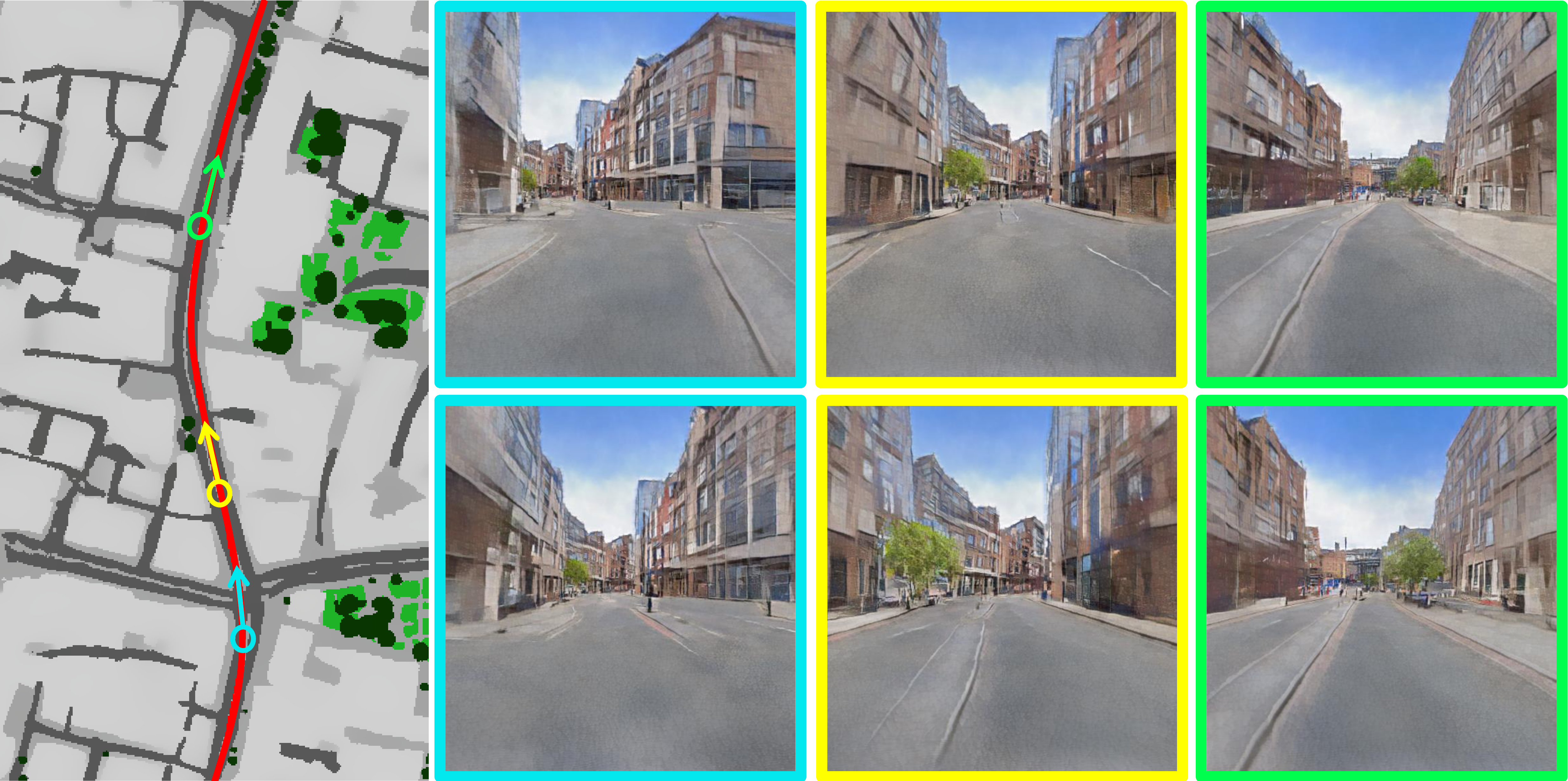} 
    \vspace{-1.5em}
    \caption{
    \textbf{Traversable and consistent 3D city rendering.}
    We render multiple views on a trajectory (red line) at three different locations (blue, yellow, and green) with small camera movements (marked as the arrow direction).
    The results show a strong cross-view consistency with a coherent global style.
    }
    % \vspace \figmargin
    % \vspace{-1.5mm}
    \label{fig:trajectory}
\end{figure} 

\vspace{-.5mm}
\subsection{Texturization via Neural Rendering}
\vspace{-1mm}
As we are the first attempt toward infinite-scale 3D environment generation jointly using 2D and 3D data, there is no existing baseline method for a fair comparison. 
Therefore, we compare with GSN~\cite{devries2021unconstrained}, a model trained on trajectories of images with camera poses, to illustrate the advantages of making use of 3D data given existing techniques.
In Figure~\ref{fig:texturization} and Figure~\ref{fig:trajectory}, we show the visual results of the images rendered on our synthesized-and-completed voxel world.
The results show a robust 3D consistency of the 3D structure, as our underlying variables and rendering mechanisms are 3D grounded.
In comparison, GSN~\cite{devries2021unconstrained} not only fails to learn the appearance of the city, but its latent space also fails to understand the 3D information due to lacking constrained and continuous camera trajectories to guide its formation.
It only creates shape deformations and 2D translations while traversing in its latent space.

Quantitatively, we further compare our method with GSN using FID~\cite{fid} and KID~\cite{binkowski2018demystifying} scores.
For GSN, the images are randomly sampled from its latent space.
For InfiniCity, we randomly sample the style latent vectors and \textit{valid} (see Section~\ref{subsec:neural}) camera poses.
% , as the images from the same trajectory share substantially overlapped content with a low variation. 
% 
% For InfiniCity, we first create a continuous camera trajectory in each environment forming a traverse video within the city, then we sub-sample the trajectory to a sparser set of points, each of the camera locations is one voxel apart from its nearest neighbor.
% 
We sample 2{,}048 images from both methods. 
The two sets of images are separately compared with a set of 2{,}048 street-view images randomly sampled from the HoliCity dataset.
The quantitative evaluation shows InfiniCity substantially outperforms GSN in both FID and KID evaluations.
% despite such an evaluation paradigm being disadvantageous to InfiniCity, as consecutive frames are sharing a certain level of content, causing both metrics to penalize the score due to the lower distributional diversity.
% 
Such a result further demonstrates our three-stage approach better captures the geometry and appearance of the city scene and is more suitable for modeling unconstrainedly large environments.

% We randomly samples images from GSN within the latent space
% \paragraph{Visual: texurization}
% \paragraph{Quantitative: FID between GSN and ours}

\vspace{-1mm}
\section{Conclusion}
\label{sec:conclusion}
\vspace{-1mm}
In this work, we propose InfiniCity, a novel framework for unbounded 3D environment synthesis. 
We demonstrate that each stage of the framework produces high-quality and high-diversity results, and together create plausible, traversable, easily editable structures at an \textit{infinite scale}.

With these exciting results, we observe that the quality of the final rendering is bounded by the neural rendering.
As neural rendering is still at its early stage with rapid revolutions, many of the convergence and efficiency problems~\cite{sun2022direct,yu2021plenoctrees,takikawa2021nglod} are being recently addressed, and we expect that our neural rendering quality will substantially improve as our understanding of such a technique is deepened.
% 
% 
% % Probably too detailed, I don't expect CVPR reviewers to understand it...
% 
% This is due to the nature of neural rendering, whose learning heavily relies on cross-view consistency.
% 
% However, as the synthetic voxel worlds do not have per-view annotated ground-truth image, where we utilize the pseudo-ground-truth to alleviate such missing information, the pseudo-ground-truths do not maintain correct/perfect cross-view consistency, therefore heavily impacting the convergence of the neural rendering.
% % 
% Fortunately, this is mainly the bottleneck caused by the optimization-based neural rendering, while more and more neural-rendering-inspired generators as being proposed, we expect future studies on transferable neural-rendering generators can lift such limitations by learning to 

%\hubert{What else? I don't find other things that are okay to reveal (e.g., generative-based neural rendering, solving cross-view consistency, unsupervised setting on landscape data...). Maybe add using the diffusion model as future work XD?}

\clearpage
%%%%%%%%% REFERENCES
{\small
\bibliographystyle{ieee_fullname}
\bibliography{main.bbl}
}

\end{document}